\documentclass[10pt,twocolumn,letterpaper]{article}

\usepackage[pagenumbers]{cvpr} 
\usepackage{url}
\usepackage{graphicx}
\usepackage{tabularx}       
\usepackage{comment}
\usepackage{makecell}
\usepackage{pifont}
\usepackage{booktabs}
\usepackage{multirow}

\usepackage[misc]{ifsym}
\usepackage{afterpage}
\usepackage{setspace}
\usepackage{breakcites}
\usepackage{seqsplit}

\usepackage{times}
\usepackage{epsfig}
\usepackage{graphicx}
\usepackage{amsmath}
\usepackage{amssymb}
\usepackage{float}
\usepackage[symbol]{footmisc}
\usepackage{colortbl}
\usepackage{amsmath}
\usepackage{stmaryrd}
\usepackage{CJKutf8}

\usepackage{bm}
\usepackage[accsupp]{axessibility} 

\definecolor{lightblue}{rgb}{0.9, 0.92, 1}
\definecolor{lightorange}{rgb}{1, 0.95, 0.9}
\definecolor{lightpink}{rgb}{1, 0.9, 0.92}
\definecolor{lightgreen}{rgb}{0.92, 1, 0.92}
\definecolor{lightred}{rgb}{1, 0.85, 0.85}

\newcommand{\colorrect}[1]{\textcolor{#1}{\ding{110}}}

\newcommand{\myparagraph}[1]{\vspace{0.2mm}\noindent\textbf{#1}}

\usepackage{lipsum}
\usepackage[dvipsnames]{xcolor}

\definecolor{cvprblue}{rgb}{0.21,0.49,0.74}
\usepackage[pagebackref,breaklinks,colorlinks,citecolor=cvprblue]{hyperref}

\def\paperID{1967} 
\def\confName{CVPR}
\def\confYear{2025}

\title{LLaVA-SLT: Visual Language Tuning for Sign Language Translation}

\vspace{-8pt}
\author{
Han Liang$^{1,2}$\footnotemark[1]\hspace{2.3em}
Chengyu Huang$^{1}$\footnotemark[1]\hspace{2.3em} 
Yuecheng Xu$^{1}$\hspace{2.3em} 
Cheng Tang$^{1}$\hspace{2.3em}
Weicai Ye$^{3}$\vspace{0.3em} \\
Juze Zhang$^{1}$\hspace{2.3em} 
Xin Chen$^{2}$\hspace{2.3em} 
Jingyi Yu$^{1}$\hspace{2.3em} 
Lan Xu$^{1}$ \vspace{0.5em} \\
$^{1}$\textit{ShanghaiTech University} \hspace{1.3em} 
$^{2}$\textit{ByteDance}\hspace{1.3em} 
$^{3}$\textit{Zhejiang University}
}

\vspace{-16pt}

\begin{document}
\maketitle
\footnotetext[1]{Equal contribution}


\begin{abstract}

In the realm of Sign Language Translation (SLT), reliance on costly gloss-annotated datasets has posed a significant barrier. Recent advancements in gloss-free SLT methods have shown promise, yet they often largely lag behind gloss-based approaches in terms of translation accuracy. To narrow this performance gap, we introduce LLaVA-SLT, a pioneering Large Multimodal Model (LMM) framework designed to leverage the power of Large Language Models (LLMs) through effectively learned visual language embeddings. Our model is trained through a trilogy.
First, we propose linguistic continued pretraining. We scale up the LLM and adapt it to the sign language domain using an extensive corpus dataset, effectively enhancing its textual linguistic knowledge about sign language. 
Then, we adopt visual contrastive pretraining to align the visual encoder with a large-scale pretrained text encoder. 
We propose hierarchical visual encoder that learns a robust word-level intermediate representation that is compatible with LLM token embeddings.
Finally, we propose visual language tuning. 
We freeze pretrained models and employ a lightweight trainable MLP connector. It efficiently maps the pretrained visual language embeddings into the LLM token embedding space, enabling downstream SLT task. 
Our comprehensive experiments demonstrate that LLaVA-SLT outperforms the state-of-the-art methods.
By using extra annotation-free data, it even closes to the gloss-based accuracy.

\end{abstract}

\section{Introduction}
\label{sec:intro}

\begin{figure}[tbp] 
	\centering  
	\includegraphics[width=1.\linewidth]{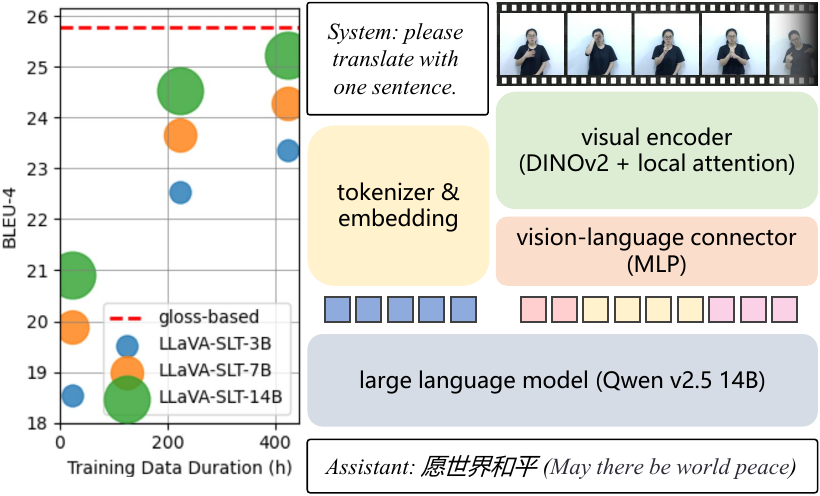} 
	\caption{LLaVA-SLT is a highly scalable LMM framework tamed for sign language translation, capable of boosting the performance with expanding multimodal data.}
	\label{fig:teaser} 
\end{figure}

The advancement of language foundation models should benefit everyone.
Sign language, as the primary mode of communication for hard-of-hearing individuals, facilitates their self-expression.
However, the complexity of learning sign language limits communication between hearing and hard-of-hearing people.
To bridge this communication gap, developing automatic sign language translation (SLT) techniques has significant social value.

Drawing on the availability of high-quality sign language datasets~\cite{zhou2021improving,camgoz2018neural}  meticulously annotated with glosses, the written sequential depiction of signs, previous approaches have achieved impressive performance~\cite{camgoz2020sign,camgoz2018neural,li2020tspnet,zhou2021improving,chen2022simple,STMC_MM,chen2022simple,chen2022two-stream,ye2023cross,yao2023sign,zhang2023sltunet}.
Despite their utility, the creation of gloss-annotated datasets is notably resource-intensive and time-consuming, presenting a significant barrier to scalability and rapid development in the field.
In response to these challenges, there has been a paradigm shift towards developing gloss-free SLT methods~\cite{zhou2023gloss,lin2023gloss}. 
These approaches aim to eliminate the dependency on gloss annotations by leveraging more generalized datasets~\cite{uthus2024youtube,duarte2021how2sign} and pretraining strategies~\cite{zhou2023gloss,jiang2024signclip}. 
However, gloss-free methods must contend with the inherent grammatical and vocabulary distinctions of sign languages, which are often poorly represented in general datasets. 
Recent advances aim to address it by utilizing the powerful linguistic capability of Large Language Models (LLMs) with various pretrained visual representations.
SignLLM~\cite{gong2024llms} transforms sign videos into discrete codes by vector quantization to improve the readability of LLMs.
FLa-LLM~\cite{chen2024factorized} employs lightweight video-grounded text generation to supervise the visual encoder.
Sign2GPT~\cite{wong2024sign2gpt} employs hand-crafted pseudo-glosses to encourage word-level alignment.
Meanwhile, SignCL~\cite{ye2024improving} proposes a sign contrastive loss to reduce representation density within dense visual sequences.
These regularization techniques constrain the visual features effectively.
However, they overlook the multimodal aspect of sign language knowledge, including distinct grammatical structures and vocabulary in both textual and visual modalities, which yields a performance gap with gloss-based methods.
On the other hand, recent advancements in Large Multimodal Models (LMMs) have ushered in new possibilities for SLT.
LMMs demonstrate cross-modal abilities across diverse inputs, including image~\cite{liu2024visual}, audio~\cite{zhang2023video}, 3D~\cite{li2024llava, chen2024ll3da,liang2024omg}, and video data~\cite{lin2023video, zhang2023video, zhang2024video}, potentially enhancing SLT. 
Yet, integrating LMMs with SLT poses unique challenges. General LMMs struggle to interpret the distinct grammatical structure of sign language, hindering translation accuracy, and require effective learning of visual language embeddings that align with LLM token embeddings.

In this paper, we present \textit{LLaVA-SLT}, a novel large multimodal model tamed for sign language translation.
Our key idea is to develop a scalable and self-compatible multimodal framework that enables continuously enhanced capabilities with expanding multimodal data (see Fig.~\ref{fig:teaser}).
For the textual modality, we perform \textbf{Linguistic Continued Pretraining} to adapt general-purpose LLMs trained on natural language to the sign language domain. 
We collect the CSL-Corpus dataset, which consists of various sign language-related corpus resources, including gloss-text pairs, books, and webpages. We then scale up the base LLM with exponentially increasing parameters, ranging from 3B to 14B, and perform generative pretraining in a standard ``next token prediction" manner.
For the visual modality, we propose \textbf{Visual Contrastive Pretraining}, utilizing contrastive learning techniques and a large-scale pretrained text encoder to pretrain the visual encoder to learn robust visual language representation. 
We propose a hierarchical architecture for the visual encoder to learn isolated local semantics within sign language videos, termed visual language, which is validated to better align with LLM token embeddings. To enhance robustness, we employ both outer contrastive loss to align with the corresponding natural language sentence and inner contrastive loss to regularize visual language embeddings, as proposed in SignCL~\cite{ye2024improving}.
Finally, we introduce \textbf{Visual Language Tuning}, employing a lightweight vision-language connector, such as a Multilayer Perceptron (MLP), mapping the learned visual language embeddings to the LLM token embedding space in an end-to-end manner, efficiently enabling downstream SLT task.

As a result, our LLaVA-SLT approach achieves
compelling gloss-free SLT performance, significantly outperforming prior arts. 
Furthermore, we conduct a systematic experimental analysis to validate the robustness and effectiveness of our framework. We hope that it will serve as a solid baseline to bring SLT into the LMM era.

To summarize, our main contributions include: 
\begin{itemize} 
\setlength\itemsep{0em}
    \item We introduce LLaVA-SLT, a LMM framework for challenging SLT task, which achieves state-of-the-art gloss-free SLT performance.
   
    \item We propose Linguistic Continued Pretraining using textual sign language corpus to effectively improve the adaptation of LLMs to the sign language domain.
   
    \item We propose a Hierarchical Visual Encoder that facilitates learning robust visual language embeddings during pretraining, along with an efficient tuning strategy that maps these pretrained embeddings into LLM token embedding space.

\end{itemize}

\section{Related Work}
\label{sec:related}

\begin{figure*}[tbp] 
	\centering  
	\includegraphics[width=1\linewidth]{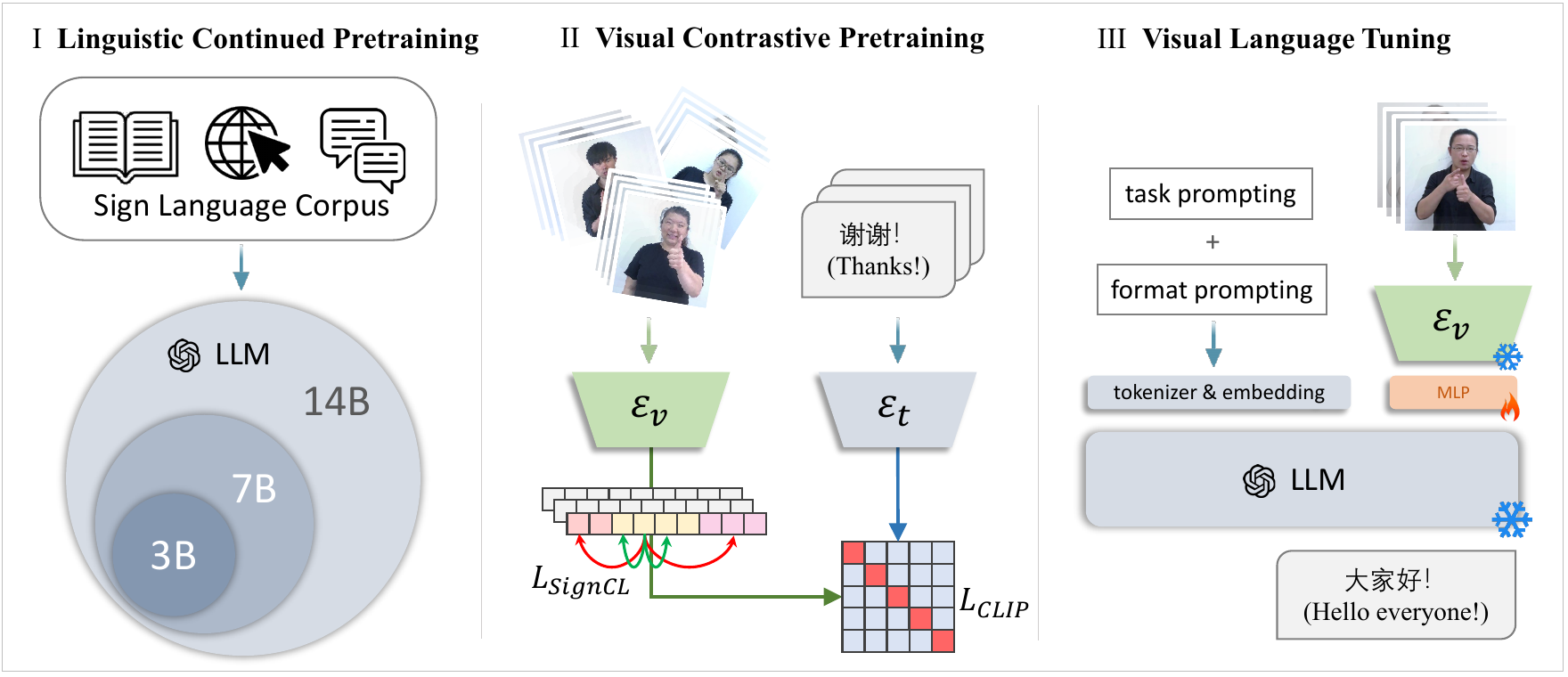} 
	\caption{\textbf{Method overview}. 
 We train LLaVA-SLT in three stages. First, we collect sign language corpus and scale up the LLMs to enhance the linguistc capabilities in sign language (Sec.~\ref{sec:3.1}).
 Subsequently, we employ a large-scale pretrained text encoder to supervise the visual encoder using both inner and outer contrastive losses (Sec.~\ref{sec:3.2}).
 Finally, we connect the pretrained visual and language models via a lightweight MLP connector, equipped with an effective prompting strategy, which efficiently enables downstream SLT task (Sec.~\ref{sec:3.3}).
 } 
	\label{fig:overview} 
\end{figure*}

\vspace{-5pt}
\myparagraph{Sign Language Translation.}
Sign Language Translation (SLT) focuses on converting sign and spoken languages, utilizing various isolated sign language datasets \cite{wang2016isolated, zhang2016chinese, joze2018ms-asl, imashev2020dataset, sridhar2020include, li2020word-level, sincand2020autsl, albanie2020bsl-1k, desai2024asl-citizen}. Recent advancements have moved from Isolated Sign Language Recognition (ISLR) \cite{albanie2020bsl-1k, tunga2021pose-based, li2020transferring, hu2021signbert} to Continuous Sign Language Recognition (CSLR), leading to new continuous datasets \cite{zhou2021improving, uthus2024youtubeasl, forster2014extensions, camgoz2018neural, vonagris2010signum, huang2018video-based, duarte2021how2sign}.
SLT methods can be categorized into gloss-based and gloss-free approaches:
Gloss-based methods use Convolutional Neural Networks (CNNs) \cite{chen2022simple, hu2023self-emphasizing, min2021visual} for feature extraction, followed by modeling with RNNs \cite{camgoz2018neural, ko2019neural}, LSTMs \cite{hu2023continuous, cui2019deep}, or Transformers \cite{voskou2021stochastic, yin2020better}. Some methods utilize keypoint sequences \cite{zhou2021spatial-temporal, papadimitriou2020multimodal} or generate heatmaps \cite{chen2022two-stream, chen2024signvtcl}. Features are decoded using CTC \cite{cheng2020fully, zhou2021spatial-temporal, min2021visual} or HMMs \cite{koller2017re-sign, gao2004chinese, koller2016deep}. However, video processing can be slow and storage-intensive, limiting practicality. Some methods enhance performance by leveraging Sign Language Recognition (SLR) results \cite{zhou2021spatial-temporal, zuo2023natural}, while others use Conditional Variational Autoencoders and Transformers for direct sign-to-text translation \cite{camgoz2018neural, zhao2024conditional, zuo2024improving}.

A key limitation of gloss-based sign language translation (SLT) is the labor-intensive gloss annotation process, which restricts scalability and results in small datasets, leading to insufficient training data. To address this, gloss-free methods \cite{camgoz2020sign, li2020tspnet} have emerged. For instance, UniGloR employs self-supervised fine-tuning, while SignVQNet utilizes sign pose quantization. Other studies \cite{chen2024factorized, gong2024llms} leverage Large Language Models (LLMs) and pseudo-translation tasks \cite{zheng2023cvt-slr}. Innovations like GASLT \cite{yin2023gloss} introduce gloss-attention, and CSGCR \cite{zhao2021conditional} enhances accuracy with word verification. 
SignLLM~\cite{gong2024llms} uses vector-quantization technique to convert sign videos into discrete sign tokens.
Sign2GPT \cite{wong2024sign2gpt} uses large-scale pre-trained visual and language models, while GFSLT-VLP \cite{zhou2023gloss-free} combines contrastive language-image pre-training with masked self-supervised learning to improve performance. SignCL \cite{ye2024improving} introduces an inner contrastive term to regularize visual embeddings by reducing representation density. Despite these advancements, gloss-free SLT methods still face a significant performance gap with gloss-based.

\myparagraph{Large Multimodal Models.}
The development of Large Multimodal Models (LMMs) has significantly integrated various modalities, including image~\cite{liu2024visual,liu2024improved}, audio~\cite{zhang2023video}, 3d~\cite{li2024llava,chen2024ll3da}, human motion~\cite{lin2024chathuman,liang2024omg}, IMU~\cite{imran2024llasa}, and video~\cite{lin2023video,zhang2023video,zhang2024video}.  Models like LLaVA \cite{liu2024visual} enhance performance through visual instruction tuning, built upon the powerful Large Language Models (LLMs).
Some advances attempt to extend it to video understanding \cite{he2024ma-lmm, li2023videochat,zhu2023minigpt-4,li2024llava-next-interleave,maaz2023video-chatgpt} and audio-text integration \cite{huang2024audiogpt,lyu2023macaw}, achieving notable progress. 
Other works like ChatPose~\cite{feng2024chatpose} and ChatHuman~\cite{lin2024chathuman} taming LMM for human image and motion reasoning, offer a new way to tackle human motion understanding.
However, they still focus on general-purpose motion reasoning, thus challenges remain in addressing the diversity of sign languages and subtle movement differences.
Recently, Sign2GPT \cite{wong2024sign2gpt} employs hand-crafted pseudo-glosses to encourage word-level alignment, aiming to tackle these challenges.
However, it still strugles to learn a robust visual sign language embedding compared to gloss-based methods.

\section{Approach}

Sign language is inherently multimodal. Our key idea is to design a highly scalable framework that can efficiently exploit the rich multimodal data of sign language.
To this end, we adapt a Large Multimodal Model (LMM) framework for sign language domain especially the SLT task.
We train our model through a trilogy.
First, for the textual modality of sign language (\cref{fig:overview} left), we collect a large-scale corpus dataset focusing on sign language topics and perform continued pretraining on open-source LLMs (\cref{sec:3.1}). 
For the video modality of sign language (\cref{fig:overview} middle), we employ contrastive learning techniques along with a large-scale pretrained text encoder to supervise the video-to-language encoder (\cref{sec:3.2}).
Finally, we finetune a lightweight MLP connector in an end-to-end fashion (\cref{fig:overview} right). 
This connector maps the pretrained visual features into the language space, efficiently bridging the gap between the well-trained vision and language models to perform the downstream SLT task (\cref{sec:3.3}).

\subsection{Linguistic Continued Pretraining} 
\label{sec:3.1}
To deepen the expertise of LLMs in sign language, especially the related knowledge and linguistic capability, we propose linguistic continued pretraining. That is to further train the powerful general-purpose LLMs with sign language corpus data in a generative pretraining manner.

\begin{figure}[tbp] 
	\centering  
	\includegraphics[width=1.01\linewidth]{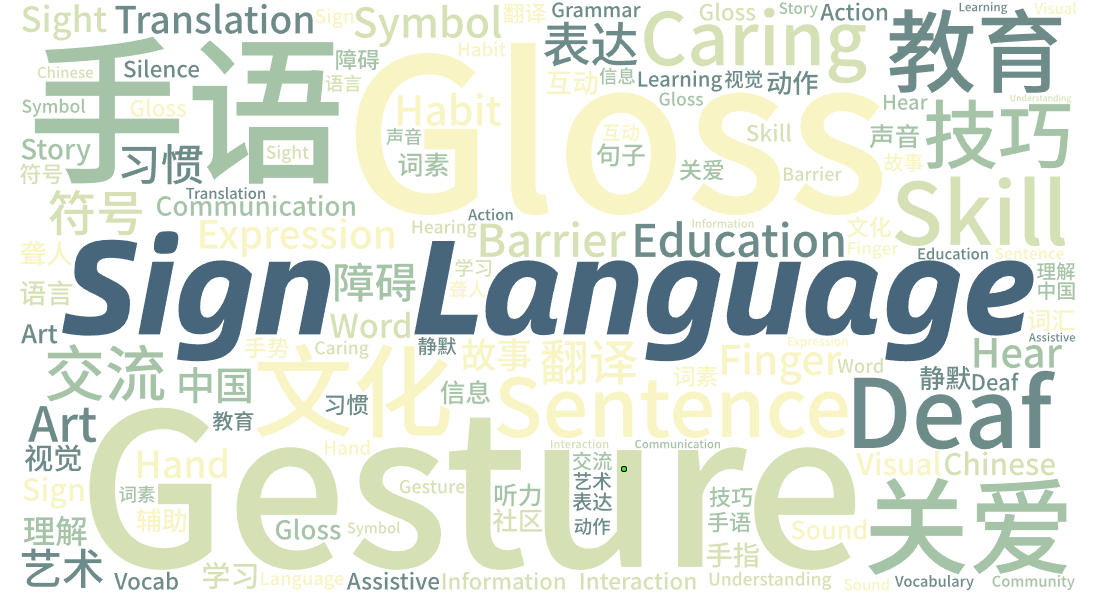} 
	\caption{Wordcloud illustration of the \textbf{CSL-Corpus} dataset. 
 } 
	\label{fig:CSL-Corpus} 
    \vspace{5pt}
\end{figure}

\myparagraph{CSL-Corpus dataset.}
To this end, we collect CSL-Corpus, a large-scale Chinese sign language corpus dataset (see Fig.~\ref{fig:CSL-Corpus}). It encompasses a wide range of resources including professional books, webpages, and gloss-text pairs. 
It is meticulously curated to facilitate the direct acquisition of CSL vocabulary and grammatical structures by LLMs, enabling learning from both instructional materials and textual representations of sign language itself.
The dataset includes a large-scale collection of over 20k textual sign language instances consisting of glosses and the corresponding texts. This data is supplemented with materials gathered from the internet, such as webpage entries and books about sign language, to form a comprehensive sign language corpus with rich textual knowledge. The dataset statistics are listed in Tab.~\ref{tab:corpus}. For more details, please refer to the supplements.

\myparagraph{Scaling base LLM.}
Recent studies on LMMs~\cite{zhao2023svit,liu2024improved} have shown that scaling up the base LLM effectively improves performance on downstream tasks. To investigate whether this scaling benefit also applies to the SLT task, we conduct an empirical study using multiple LLMs of varying sizes. Specifically, we adopt decoder-only transformer-based models due to their exceptional scalability and proven cross-modal capabilities~\cite{liu2024improved,zhang2023video,chen2024ll3da,imran2024llasa,han2024onellm}. Among the available models, we employ \emph{Qwen-2.5-Instruct}~\cite{yang2024qwen2,hui2024qwen2} as CSL base LLM for its state-of-the-art performance in Chinese domain. To systematically assess the impact of model size on SLT performance, we select models logarithmically in three different sizes, ranging from 3 billion to 14 billion parameters.

\myparagraph{Training.}
We train our LLMs in a standard ``next token prediction" manner.
Given the trainable weights $\mathbf{\theta}$ and an $|\mathbf{c}|$-length unsupervised corpus tokens $\mathbf{c}$, the auto-regressive loss is calculated as:
\begin{equation}
    \mathcal{L}_{\text{AR}}\left(\theta\right) = - \sum_{i=1}^{|\mathbf{c}|} \log P\left(\mathbf{c}_{\left[i\right]} \vert \mathbf{c}_{\left[i-k, \cdots, i-1\right]}, \mathbf{\theta} \right),
    \label{eq:ce_loss}
\end{equation}
where $k$ is the size of the sliding context window.

During training, for the paired gloss-text data, their order is randomly permuted to enable learning both the intrinsic grammatical structure of sign language and its relation with natural language. For the other corpora, we employ a sliding context window to sample from the entire text. 
To improve memory efficiency to train larger models, we incorporate Low-Rank Adapters (LoRA)~\cite{hu2021lora}, a widely used parameter-efficient finetuning method. In addition, controlling trainable parameters through the hyperparameter rank $\mathbf{\gamma}$ simplifies the balance between learning and forgetting~\cite{biderman2024lora}.

In our experiments, all models are trained for 1 epoch with a batch size of $64$ using the AdamW~\cite{loshchilov2017decoupled} optimizer with a weight decay of $0$, a maximum learning rate of $2e^{-4}$, and a one-cycle cosine learning rate scheduler. LoRA rank $\mathbf{\gamma}=8$ and alpha scaling factor $\mathbf{\alpha}=16$.
The models are trained using Pytorch with ZeRO~\cite{rajbhandari2020zero} memory redundancy strategy on $8\times$ NVIDIA A100 GPUs.

\begin{table}[tbp]
\centering
\setlength{\tabcolsep}{3pt}
\resizebox{0.85\linewidth}{!}{
    \begin{tabular}{lcccc} 
        \toprule
        \textbf{Source} & \textbf{Language} & \textbf{Tokens} & \textbf{Vocab} & \textbf{Topic} \\  
        \midrule
        \rowcolor{gray!20} gloss-text pairs & Chinese & 321k & 3,600 & CSL \\
        books & Chinese & 261k & 4,686 & CSL \\
        \rowcolor{gray!20} webpages & English/Chinese & 155k & 4,010 & CSL \\
        \bottomrule
    \end{tabular}
}
\caption{Statistics of the \textbf{CSL-Corpus} dataset.}
\label{tab:corpus}
\end{table}

\begin{table}[t]
\centering
\resizebox{\columnwidth}{!}{
    \begin{tabular}{lccccccc} 
        \toprule
        \textbf{Dataset} & \textbf{Language} & \textbf{Videos} & \textbf{Duration} & \textbf{Vocab} & \textbf{Gloss} & \textbf{Source}  \\
        \midrule
        \rowcolor{gray!20} PHOENIX-2014~\cite{forster2014extensions}  & DGS & 6,841 & 8.9h & 1k     &   $\times$    & TV   \\
        PHOENIX-2014T~\cite{camgoz2018neural} & DGS & 8,257 & 11h  & 3k     &  \checkmark   & TV  \\
         \rowcolor{gray!20} CSL-Daily~\cite{zhou2021improving} & CSL & 20,654 & 22.8h  & 2k    &  \checkmark   & Lab \\
        \midrule
        \textbf{CSL-400h (Ours)} & CSL & \textbf{264,461}  & \textbf{401h}  & \textbf{5k}    & $\times$  & Lab \\
        \bottomrule
    \end{tabular}
}
\caption{\textbf{CSL-400h dataset.} Thanks to gloss-free, our dataset is currently the largest video dataset in Chinese sign language, with a duration that is 10$\times$ greater than existing dataset.}
\label{tab:dataset}
\vspace{5pt}
\end{table}

\subsection{Visual Contrastive Pretraining} 
\label{sec:3.2}
Previous works~\cite{zhou2023gloss,ye2024improving} have demonstrated that contrastive vision-language pretraining improves the robustness of the vision encoder and mitigates overfitting issues in the end-to-end training.
At this stage, we adopt the standard CLIP~\cite{radford2021learning} training paradigm to align our visual encoder $\mathcal{E}_v$ with a large-scale pretrained text encoder $\mathcal{E}_t$.

\begin{figure}[tbp] 
	\centering  
 
	\includegraphics[width=0.99\linewidth]{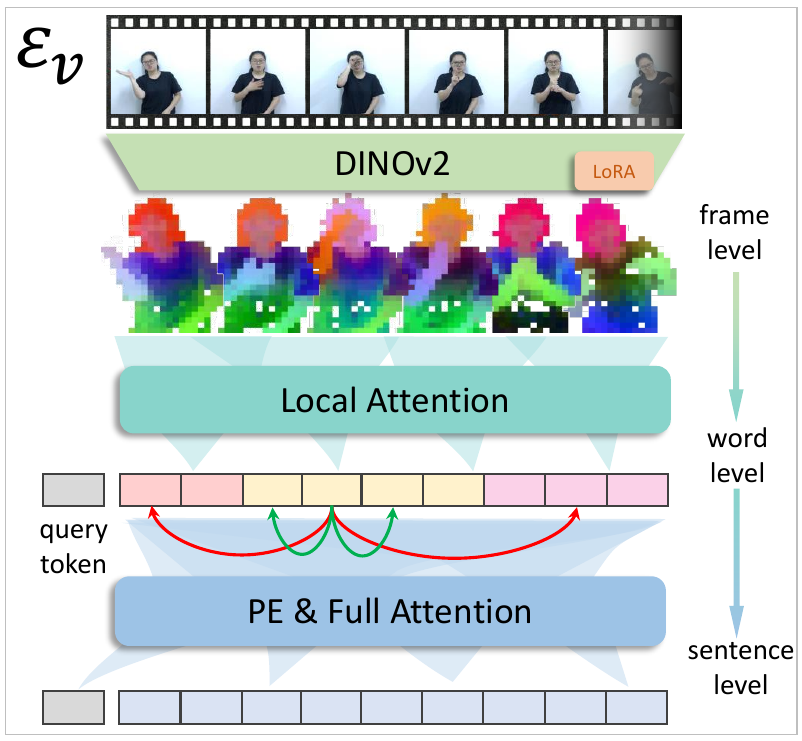} 
	\caption{\textbf{Hierarchical visual architecture.}  The visual representation is explicitly separated into three levels: frame level, word level, and sentence level.
 } 
	\label{fig:control} 

\end{figure}

\myparagraph{Hierarchical visual encoder.}
Given the relatively independent multi-submotion structure of sign language, we design our visual encoder with a hierarchical architecture. It allows the introducing of appropriate inductive biases at different feature levels, including frame-level, word-level, and full sentence-level features, as illustrated in Fig.~\ref{fig:control}.

Specifically, for the frame level, we utilize the large-scale pretrained image encoder, DINOv2~\cite{oquab2023dinov2}, to provide robust visual features of the human body from individual frames. To adapt the general features of DINOv2 to capture critical aspects such as gestures, mouth shapes, and facial expressions, we finetune the model with LoRA similar to \cite{wong2024sign2gpt}. 
For the word level, to extract independent ``words" from local subsequences of frame-level features, we employ a local-attention transformer with rotary positional embeddings~\cite{su2024roformer}. The receptive field is confined to a local window, which is minimized to cover only each individual ``word" in the dataset. SignCL loss~\cite{ye2024improving} is applied here to regularize the word embedding space.
And finally, we treat the sequence of word-level features as a full sentence and encode it into a query token using the standard transformer encoder, which is used to align with the corresponding natural language embedding.

\myparagraph{Scaling model and resolution.}
Inspired by the performance gain in LLaVA-1.5~\cite{liu2024improved} by scaling visual encoder,
we investigate the scaling benefit in the visual modality of the sign language domain. 
We scale up not only the model size but also the input image resolution from $224^2$ to $336^2$, to allow the model to clearly ``see" the details of human body movements.

\myparagraph{Scaling data.}
Besides, we also explore the scaling law on the data side, expanding data in exponential space up to $10\times$ the existing dataset.
Thanks to gloss-free, the burdensome annotation process is eliminated, facilitating more cost-effective and time-efficient scaling.
Specifically, we collect a large-scale video-text paired CSL dataset, named CSL-400h.
Our dataset consists of 264,461 videos, up to 401 hours, along with their corresponding natural language version. Statistics of the dataset can be found in Tab.~\ref{tab:dataset}.
For more details, please refer to the supplements.



\myparagraph{Training.}
The visual encoder is trained using the standard language-supervised contrastive loss introduced by CLIP~\cite{radford2021learning}. In addition, as recommended by SignCL~\cite{ye2024improving}, we also apply inner contrastive loss that serves as a regularization term to regularize the embedding space.
The total contrastive learning loss is calculated as:
\begin{equation}
    \mathcal{L}_{\text{CL}} = \mathcal{L}_{\text{CLIP}} + \lambda \mathcal{L}_{\text{SignCL}},
    \label{eq:ce2_loss}
\end{equation}
where $\lambda$ is a hyperparameter to balance the inner and outer contrastive loss.

In our experiments, we employ the \emph{ViT-B/14-distilled} variant of DINOv2~\cite{oquab2023dinov2} as the backbone and use its $\left[ cls \right]$ token as the full image feature. It is tuned by LoRA with a rank $\gamma=8$ and an alpha scaling factor $\alpha=16$. 
The layer depth of the local-attention transformer is set to 4. 
The nearest-neighbor downsample with step 4 is applied to the word-level features along with the SignCL loss of strength $\lambda=1e^{-2}$.
The layer depth of full-attention transformer is set to 8 and the dimension of CLIP co-embedding is set to 768.
We employ the pretrained text encoder from \emph{mBART}~\cite{liu2020multilingual} due to its multilingual capability. 
We train all models for $200$ epochs with a batch size of $128$ using the AdamW optimizer with a weight decay of $1e^{-5}$, a maximum learning rate of $2e^{-4}$, and a one-cycle cosine learning rate scheduler. 

\subsection{Visual Language Tuning} 
\label{sec:3.3}
Given the pretrained vision and language models specialized in sign language, we draw inspiration from the recent advances in visual instruction tuning techniques~\cite{liu2024visual,liu2024improved} to introduce Visual Language Tuning (VLT). VLT facilitates the seamless integration of a pretrained visual language model with a textual language model, thereby enhancing the model's capability on downstream tasks like SLT.

\myparagraph{Vision-language connector.}
Like other visual tuning methods~\cite{liu2024visual}, VLT freezes the parameters of both the visual encoder and the LLM.
It incorporates a lightweight vision-language connector, i.e. a compact neural network, to map the pretrained vision features into the LLM token embedding space.
We choose a two-layer MLP as the connector as LLaVA-1.5~\cite{liu2024visual} proposed.
The difference is that we explicitly map the word-level visual tokens, i.e. visual language, into the LLM token embedding space, instead of the widely adopted image tokens of the penultimate layer used in the general-purpose LMMs~\cite{liu2024visual,liu2024improved,lin2023video,li2024llava}.
We find that this achieves significantly enhanced performance, which indicates the effectiveness of visual language learning with hierarchical sign language representations.

\myparagraph{Response prompting.}
To address the ambiguity in the response content and format for the specific translation task within a predefined instruction template of LLMs, we propose response prompting which includes task prompting and format prompting, as illustrated in Tab.~\ref{subtab:format_prompts_diff}. 
It uses two simple sentences to explicitly define the task and prescribe a clear format for the responses, respectively.
These prompts are appended at the end of the system prompt during both training and inference.
We then insert the output of the MLP connector into the user's input part of the template, i.e. concatenating them with text token embeddings as the overall input of LLMs, as illustrated in Fig.~\ref{fig:overview}.

\myparagraph{Training.}
We train the connector using the same cross-entropy loss as in Eqn.~\ref{eq:ce_loss}, which is only applied to the response part.
In our experiments, it is trained with a batch size of $64$ using AdamW optimizer with a weight decay of $1e^{-3}$, a maximum learning rate of $1e^{-3}$, and a one-cycle cosine learning rate scheduler. 
All models are trained for several epochs until their validation performance stops improving.
Furthermore, we unfreeze the visual encoder and LLM (LoRA if used) and perform further full-tuning with a small learning rate of $1e^{-5}$.
We find this further improves the performance slightly.
\section{Experimental Evaluation}
\label{sec:exp}

\begin{table}[tbp]
\centering

\resizebox{0.47\textwidth}{!}{
    \begin{tabular}{lccccc}
        \toprule
        \multirow{2}{*}{\textbf{Method}} & \multicolumn{5}{c}{\textbf{Test Set}} \\
        \cmidrule(lr){2-6}
         & \textbf{BLEU1} & \textbf{BLEU2} & \textbf{BLEU3} & \textbf{BLEU4} & \textbf{ROUGE}  \\
         \midrule
          \rowcolor{gray!40} \multicolumn{6}{c}{\textbf{Gloss-based}}\\
          \midrule
        SL-Transformer \citep{camgoz2020sign}   & $37.38$ & $24.36$ & $16.55$ & $11.79$ & $36.74$\\
        BN-TIN-Transf.+BT  \citep{zhou2021improving}   &  $51.42$ & $37.26$ & $27.76$ & $21.34$ & $49.31$\\
        MMTLB \citep{chen2022simple}   & $53.31$ & $40.41$ & $30.87$ & $23.92$ & $53.25$\\
        SLTU$_{\texttt{NET}}$  \citep{zhang2023sltunet}  & $54.98$ & $41.44$ & $31.84$ & $25.01$& $54.08$\\
        TwoStream-SLT \citep{chen2022two-stream}   & $55.44$ & $42.59$ & $32.87$ & $25.79$ & $55.72$\\
        
         \midrule
         \rowcolor{gray!40}
         \multicolumn{6}{c}{\textbf{Gloss-free}}\\
        \midrule
        GASLT \citep{yin2023gloss}      & $19.90$ & $9.94$  & $5.98$  & $4.07$  & $20.35$ \\
        NSLT \citep{camgoz2018neural}  & $34.16$ & $19.57$ & $11.84$ & $7.56$  & $34.54$ \\
        GFSLT \citep{zhou2023gloss}     & $37.69$ & $23.28$ & $14.93$ & $9.88$ & $35.16$ \\
        GFSLT-VLP \citep{zhou2023gloss}     & $39.37$ & $24.93$ & $16.26$ & $11.00$ & $36.44$ \\
        Fla-LLM~\cite{chen2024factorized} & $37.13$ & 25.12 & 18.38 & 14.20 & $37.25$ \\
        SignLLM~\cite{gong2024llms} & 39.55 & 28.13 & 20.07 & 15.75 & 39.91 \\
        Sign2GPT-PGP \citep{wong2024sign2gptleveraginglargelanguage} & $41.75$ & $28.73$ & $20.60$ & $15.40$ & $42.36$ \\
        GFSLT-VLP-SignCL \citep{ye2024improving}  & $\underline{47.47}$ & $\underline{32.53}$ & $\underline{22.62}$ & $\underline{16.16}$ & $\underline{48.92}$ \\
        \midrule
        \textbf{LLaVA-SLT (ours)}  & $\textbf{52.15}$ & $\textbf{36.24}$ & $\textbf{26.47}$ & $\textbf{20.42}$ & $\textbf{51.26}$ \\
        \textbf{LLaVA-SLT (w/ extra data)}  & $56.38$ & $42.76$ & $32.43$ & $25.23$ & $56.21$ \\
        \bottomrule
    \end{tabular}
}
\caption{Comparison of test set results on the CSL-Daily. We present results for two settings: (1) with the same training set (\textbf{ours}), and (2) adding the extra CSL-400h data (\textbf{w/ extra data}).}
\label{tab:3}
\vspace{-5pt}
\end{table}

\subsection{Benchmarks}
\label{sec:4.2}

\myparagraph{Datasets.}
CSL-Daily~\cite{zhou2021improving} and Phoenix-2014T~\cite{camgoz2018neural} are utilized as the primary datasets to evaluate our proposed method. CSL-Daily is a dataset designed for translating Chinese sign language into spoken Chinese. It consists of laboratory-recorded videos that capture a wide range of everyday interactions, such as travel, family life, banking, and shopping, providing a diverse linguistic context for translation tasks. Phoenix-2014T is tailored for German sign language to German translation, sourced from televised German weather broadcasts. It serves as a benchmark for evaluating sign-to-spoken language translation in a continuous, natural setting.

\myparagraph{Metrics}
To assess translation performance, following previous works~\cite{ye2024improving,chen2022two-stream}, we employ the standard metrics BLEU (Bilingual Evaluation Understudy)~\cite{papineni2002bleu} and ROUGE-L~\cite{lin2004rouge}, both commonly used in sign language translation tasks. BLEU quantifies the degree of overlap between model-generated translations and reference texts by comparing n-grams, where higher scores indicate greater lexical similarity. ROUGE-L, on the other hand, evaluates translation quality by identifying the longest common subsequence between the generated and reference sentences, providing insight into structural alignment. For both metrics, a higher score reflects improved translation accuracy and quality.

\begin{table}[tbp]
\centering

\resizebox{0.47\textwidth}{!}{
    \begin{tabular}{lccccc}
        \toprule
        \multirow{2}{*}{\textbf{Method}} & \multicolumn{5}{c}{\textbf{Test Set}} \\
        \cmidrule(lr){2-6}
         & \textbf{BLEU1} & \textbf{BLEU2} & \textbf{BLEU3} & \textbf{BLEU4} & \textbf{ROUGE}  \\
         \midrule
          \rowcolor{gray!40} \multicolumn{6}{c}{\textbf{Gloss-based}}\\
          \midrule
            SL-Transformer \citep{camgoz2020sign}   & $46.61$ & $33.73$ & $26.19$ & $21.32$ & $-$\\
            BN-TIN-Transf.+BT  \citep{zhou2021improving}   &  $50.80$ & $37.75$ & $29.72$ & $24.32$ & $49.54$\\
            MMTLB \citep{chen2022simple}   & $53.97$ & $41.75$ & $33.84$ & $28.39$ & $52.65$\\
            SLTU$_{\texttt{NET}}$  \citep{zhang2023sltunet}  & $52.92$ & $41.76$ & $33.99$ & $28.47$ &  $52.11$\\
            TwoStream-SLT \citep{chen2022two-stream}   & $54.90$ & $42.43$ & $34.46$ & $28.95$ & $53.48$\\
            
             \midrule
             \rowcolor{gray!40}
             \multicolumn{6}{c}{\textbf{Gloss-free}}\\
            \midrule
            NSLT \citep{camgoz2018neural}   & $29.86$ & $17.52$ & $11.96$ & $9.00$ & $30.70$ \\
            TSPNet \citep{li2020tspnet}     & $36.10$ & $23.12$ & $16.88$ & $13.41$ & $34.96$ \\
            CSGCR \citep{zhao2021conditional}        & $36.71$ & $25.40$ & $18.86$ & $15.18$ & $38.85$ \\
            GASLT \citep{yin2023gloss}         & $39.07$ & $26.74$ & $21.86$ & $15.74$ & $39.86$ \\
            GFSLT \citep{zhou2023gloss}   & $41.39$ & $31.00$ & $24.20$ & $19.66$ & $40.93$ \\
            GFSLT-VLP \citep{zhou2023gloss}   & $43.71$ & $33.18$ & $26.11$ & $21.44$ & $42.49$ \\
        SignLLM~\cite{gong2024llms} & 45.21 & 34.78 & 28.05 & \underline{23.40} & 44.49 \\
        Fla-LLM~\cite{chen2024factorized} & 46.29 & 35.33 & 28.03 & 23.09 & 45.27 \\
        Sign2GPT-PGP \citep{wong2024sign2gptleveraginglargelanguage} & 49.54 & 35.96 & 28.83 & 22.52 & 48.90 \\
        GFSLT-VLP-SignCL \citep{ye2024improving}  & $\underline{49.76}$ & $\underline{36.85}$ & $\textbf{29.97}$ & $22.74$ & $\underline{49.04}$ \\
        \midrule
        \textbf{LLaVA-SLT (ours)}  & $\textbf{51.20}$ & $\textbf{37.51}$ & $\underline{29.39}$ & $\textbf{23.43}$ & $\textbf{50.44}$ \\
        \bottomrule
    \end{tabular}
}
\caption{Comparison of test set results on the Phoenix-2014T. }
\label{tab:4}
\vspace{-5pt}
\end{table}

\subsection{Results}
\label{sec:4.3}
Here we show that LLaVA-SLT achieves superior performance over previous gloss-free methods. 
We conduct comparative analysis against existing state-of-the-art models, particularly GFSLT-VLP-SignCL~\cite{ye2024improving}, which also employs contrastive learning techniques. Quantitative evaluation is presented on the CSL-Daily dataset (see Tab.~\ref{tab:3}), where \emph{\textbf{ours}} denotes our model trained using the CSL-Daily training set. We observe that our approach surpasses other gloss-free methods with a significant improvement in terms of the BLEU-4 score, even close to gloss-based methods. 
Notably, after incorporating our additional annotation-free CSL-400h dataset for enhanced training, the last row denoted by \emph{\textbf{w/ extra data}}, LLaVA-SLT surpasses gloss-based methods in BLEU-1 scores. The qualitative comparison shown in Tab.~\ref{tab:5} also illustrates the superiority. It is promising and suggests that, with the expansion of web-scale unlabeled data, gloss-free methods could potentially replace the cumbersome and labor-intensive annotation-dependent approaches. Our method establishes a solid baseline that could stimulate further research in this field.

Further validation on generalizability is conducted on the Phoenix-2014T dataset. We select \emph{LLaMA-3.1-8B-Instruct}~\cite{dubey2024llama} as the base LLM and text encoder of \emph{mBART}~\cite{liu2020multilingual} as visual supervisor due to its multilingual capability, particularly in German. Notably, we skip linguistic continued pretraining stage due to the lack of German corpus, but our model still outperforms the previous arts, as shown in Tab.~\ref{tab:4}.

\begin{table}[t!]
\centering
\scalebox{0.78}{
  \begin{tabular}{p{1.5cm} p{8.2cm} }
    \toprule
    \multicolumn{2}{l}{\bf Qualitative comparison}  \\
    \midrule
    Reference & \begin{CJK}{UTF8}{gbsn} 你冷静点，我会帮你的。\end{CJK} \\
              &(Calm down, I will help you.) \\
    SignCL~\cite{ye2024improving} & \begin{CJK}{UTF8}{gbsn} \colorbox{lightred}{我会帮助你安静下来}。\end{CJK} \\
              &(\colorbox{lightred}{I will help you calm down}.) \\
    Ours & \begin{CJK}{UTF8}{gbsn} 你冷静下来，我会帮助你的。\end{CJK} \\
              &(Calm down, I will help you.) \\
    \midrule
    Reference & \begin{CJK}{UTF8}{gbsn}上海可以说是中国的金融中心。\end{CJK} \\
              &(Shanghai can be considered Chinese financial center.) \\
    SignCL~\cite{ye2024improving} & \begin{CJK}{UTF8}{gbsn} 上海\colorbox{lightred}{人很关心国家的经济}。\end{CJK} \\
              &(Shanghai \colorbox{lightred}{residents concerns about the national economy}.) \\
    Ours & \begin{CJK}{UTF8}{gbsn} 上海是中国金融中心。\end{CJK} \\
              &(Shanghai is Chinese financial center.) \\
    \midrule
    Reference & \begin{CJK}{UTF8}{gbsn} 你女儿的房间很漂亮，都是粉色的。\end{CJK} \\
              &(Your daughter's room is very pretty, all in pink.) \\
    SignCL~\cite{ye2024improving} & \begin{CJK}{UTF8}{gbsn} 
你女儿\colorbox{lightred}{打扮得很漂亮}，全身粉色。\end{CJK}\\
              &(Your daughter\colorbox{lightred}{looks lovely, dressed}all in pink.) \\
    Ours & \begin{CJK}{UTF8}{gbsn} 女儿的房间很漂亮，全是粉红色的。\end{CJK} \\
              &(Daughter's room is very pretty, all in pink.) \\
    \bottomrule
  \end{tabular}
}
\caption{\textbf{Qualitative comparison.} Compared to GFSLT-VLP-SignCL, our method captured key words more accurately, while preserving the original semantics effectively.}  
\label{tab:5}
\end{table}

\subsection{Ablation on LLM Continued Pretraining}

Our investigation into the impact of LLM pretraining starts with a baseline model, \emph{Qwen-2.5-3B-Instruct} without pretraining. We incrementally add key factors to observe performance changes, with the same visual encoder and other configurations for fair comparison.

\myparagraph{Effect of various corpus data.}
Initially, we augment the baseline model by progressively incorporating gloss-text data and web\&books data from the CSL-Corpus dataset for pretraining. The results, shown in Tab.~\ref{tab:6scaling_ablation} \emph{Row 1} and \emph{Row 2}, demonstrate a steady performance improvement, notably with gloss-text data. This highlights the effectiveness of pretraining LLMs in sign language textual representations and domain-specific knowledge.

\myparagraph{Effect of base LLM scaling.}
The efficacy of LLMs often hinges on the model size, which is typically large and pretrained on extensive datasets. This fundamentally affects the performance of downstream tuning. Thus we scale up the baseline model from 3B to 7B and 14B, based on pretraining with CSL-Corpus data. The results, depicted in Tab.~\ref{tab:6scaling_ablation} \emph{Row 3} and \emph{Row 4}, indicate that performance consistently improves with model scaling, underscoring the general linguistic capacity of the base LLM to enhance the SLT task.

\myparagraph{Effect of base LLM choice.}
Furthermore, we test the impact of downgrading the 7B model to a previous version, Qwen-1.5~\cite{bai2023qwen}, and substituting it with other popular LLMs, such as LLaMA-3.1. As shown in Tab.~\ref{tab:6scaling_ablation} \emph{Row 5} and \emph{Row 6}, the former results in a slight performance decline, while the latter leads to a more significant drop. This can be attributed to the fact that LLaMA’s training data predominantly comprises English, lacking the extensive Chinese corpus used in Qwen.
This also suggests the importance of base LLM continued training for specific domains.

\begin{table}[t!]
\centering
\scalebox{0.78}{
\begin{tabular}{l l c | ccc}
\toprule
\multicolumn{2}{l}{Method} & LLM  & BLEU1 & BLEU4 & ROUGE\\
\midrule
\multicolumn{6}{l}{\it Keep the final visual encoder and other configurations} \\
 & \textbf{Qwen-2.5 (w/o CPT)} & 3B  & 46.56 & 16.69 & 45.59 \\
\rowcolor{lightgreen} 1) & + gloss-text data & 3B  & 48.79 & 18.09 & 46.72 \\
\rowcolor{lightgreen} 2) & + web\&books data & 3B  & 49.02 & 18.54 & 47.13 \\
\rowcolor{lightblue} 3) & + scaling & 7B  & \underline{50.94} & \underline{19.88} & \underline{49.70} \\
\rowcolor{lightblue} 4) &  + scaling++ & 14B  & \textbf{52.15} & \textbf{20.92} & \textbf{51.26} \\
\rowcolor{lightorange} 5) & - Qwen-1.5 & 7B  & 50.27 & 19.05 & 49.52 \\
\rowcolor{lightorange} 6) & - Llama-3.1 & 8B  & 48.95 & 18.39 & 48.19 \\

\bottomrule
\end{tabular}
}
\caption{
\textbf{LLM pretraining ablation} on \colorrect{lightgreen} data, \colorrect{lightblue} model size and \colorrect{lightorange} base model choice. 
}
\label{tab:6scaling_ablation}
\vspace{-5pt}
\end{table}

\begin{table}[t!]
\centering
\scalebox{0.76}{
\begin{tabular}{p{2mm} l p{7mm}p{5mm} |ccc}
\toprule
\multicolumn{2}{l}{Method} & Dino & Res.& BLEU1 & BLEU4 & ROUGE\\
\midrule
\multicolumn{7}{l}{\it Keep the final 7B LLM and other configurations} \\
 &  \textbf{LLaVA-SLT-7B} & ViT/B & 336 & 50.94 & 19.88 & 49.70 \\
 \rowcolor{lightblue} 1) & - local attention & ViT/B & 336 & 45.47 & 14.87 & 44.76 \\
\rowcolor{lightblue} 2) & - backbone size & ViT/S & 336 & 50.09 & 19.25 & 48.58 \\
\rowcolor{lightblue} 3) & - input resolution & ViT/B & 224 & 48.96 & 18.42 & 48.01 \\
\rowcolor{lightgreen} 4) & + 50\% extra data & ViT/B & 336 & \underline{53.78} & \underline{23.65} & \underline{52.83} \\
\rowcolor{lightgreen} 5) & + 100\% extra data & ViT/B & 336 & \textbf{55.18} & \textbf{24.28} & \textbf{54.21} \\
\rowcolor{lightorange} 6) & w/ Qwen supervision & ViT/B & 336 & 40.77 & 13.42 & 38.95 \\

\bottomrule
\end{tabular}
}
\caption{
\textbf{Visual pretraining ablation} on \colorrect{lightblue} visual encoder, \colorrect{lightgreen} data, and \colorrect{lightorange} text encoder choice. 
}
\label{tab:7scaling_ablation}
\end{table}

\subsection{Ablation on Visual Pretraining}
In Tab.~\ref{tab:7scaling_ablation}, we conduct an ablation study on the visual encoder pretraining. Starting from the LLaVA-SLT-7B model with the pretrained 7B LLM, we substitute key submodules while keeping the same other configurations.

\myparagraph{Evaluation on hierarchical architecture.}
To evaluate the hierarchical architecture, we remove the local-attention transformer and directly use frame-level features as input into the full-attention transformer. As shown in Tab.~\ref{tab:7scaling_ablation} \emph{Row 1}, this results in a notable performance decrease, with the BLEU-4 score dropping from 19.88 to 14.87. It indicates that the word-level intermediate representation using the local attention mechanism effectively enhances the robustness and compatibility of visual language tokens.

\myparagraph{Effect of visual scaling.}
On the scale side, we reduce the size of the DINOv2 backbone from \emph{ViT/B-14} to \emph{ViT/S-14} and decrease the input image resolution from $336^2$ to $224^{2}$. A slight decline is observed with the size decreasing, while the lower resolution leads to a more significant drop (Tab.~\ref{tab:7scaling_ablation} \emph{Row 2} vs. \emph{Row 3}). This suggests that higher resolution is crucial for tasks like SLT, where fine-grained human motion details are essential. In addition, a larger vision backbone can also positively contribute to the final performance.
We further investigate the impact of data scaling by incrementally expanding the training data from baseline CSL-Daily (approximately 20 hours) to 200 hours and 400 hours using our CSL-400h dataset. The results (Tab.~\ref{tab:7scaling_ablation} \emph{Row 4} and \emph{Row 5}) show a clear gradient boost as the data scale increases exponentially, with BLEU-1 scores increasing from 50.94 to 53.78 and 55.18, respectively.
Moreover, we replace the text encoder with the \emph{Qwen-2.5-0.5B} model to supervise the training of the visual encoder, yet surprisingly resulting in a markedly poor performance (Tab.~\ref{tab:7scaling_ablation} \emph{Row 6}). 
The results indicate that the alignment issues associated with the auto-regressive LLMs manifest a reduced efficacy, compared to the general text encoders.

\subsection{Ablation on Visual Language Tuning}
\label{sec:4.4}
In this section, we assess the technical contributions of our proposed visual language tuning method by evaluating the impact of specific technical choices on LLaVA-SLT-7B's performance, as shown in Tab.~\ref{tab:scaling_ablation}.

\myparagraph{Effect of connector choices.}
We first replace word-level features with sentence-level features (from the penultimate layer), resulting in notable performance degradation as shown in Tab.~\ref{tab:scaling_ablation} \emph{Row 1}, particularly evident in the ROUGE score dropping from 49.70 to 46.78.  This indicates that word-level visual language embeddings align more effectively with the LLM embedding space compared to sentence-level representation. It also highlights our strategy of limiting the receptive field, enabling the model to accurately capture the local semantics of sign language, rather than memorizing the entire sequence.  In addition, substituting the MLP with a linear projection leads to a significant performance drop, as shown in Tab.~\ref{tab:scaling_ablation} \emph{Row 2}, suggesting that the relation between visual language embeddings and the LLM token embeddings is not merely linear.

\myparagraph{Evaluation on response prompting.}
We further investigate the effect of our LLM prompting strategies. We incrementally remove format prompting and task prompting while keeping the other configurations same. As indicated in Tab.~\ref{tab:scaling_ablation} \emph{Row 3} and \emph{Row 4}, the removal of format prompting results in a relatively minor performance degradation. Conversely, eliminating all forms of prompting leads to a significant drop in performance.
This suggests that our prompting mechanism effectively constrains the content and format of the LLM's output, narrowing the output space and thereby enabling a more efficient search for the vision-language mappings. This is also depicted by the qualitative results shown in Tab.~\ref{subtab:format_prompts_diff}.
Furthermore, we present results without further full-tuning, in \emph{Row 5} of Tab.~\ref{tab:scaling_ablation}. This results in a slight decline in performance, indicating that the pretrained vision and language models still exhibit a minor modality gap, which is effectively mitigated through further full-tuning.

\begin{table}[t!]
\centering
\scalebox{0.8}{
\begin{tabular}{l l p{7mm}| cc c}
\toprule
\multicolumn{2}{l}{Method}  & LLM  & BLEU1 & BLEU4 & ROUGE\\
\midrule
\multicolumn{6}{l}{\it Keep the final LLM and visual encoder} \\
 & \textbf{LLaVA-SLT-7B} & 7B  & \textbf{50.94} & \textbf{19.88} & \textbf{49.70} \\
 \rowcolor{lightblue} 1) & w/ sentence-level feature & 7B   & 48.70 & 18.65 & 46.78 \\
\rowcolor{lightblue} 2) & w/ linear connector & 7B & 46.80 & 17.05 & 45.17 \\
\rowcolor{lightgreen} 3) & w/o format prompting & 7B  & 49.65 & 19.01 & 48.46 \\
\rowcolor{lightgreen} 4) & w/o prompting & 7B   & 46.13 & 16.79 & 45.80 \\
\rowcolor{lightorange} 5) & w/o additional full-tuning & 7B   & \underline{50.04} & \underline{19.40} & \underline{48.92} \\

\bottomrule
\end{tabular}
}
\caption{
\textbf{Visual tuning ablation} on \colorrect{lightblue} connector, \colorrect{lightgreen} prompt, and \colorrect{lightorange} additional full-tuning.
}
\label{tab:scaling_ablation}
\end{table}

\begin{table}[t!]
\centering
\scalebox{0.94}{
  \begin{tabular}{p{2.4cm} p{5.7cm} }
    \toprule
    \multicolumn{2}{l}{\bf Different Prompts}  \\ 
    \midrule
    Reference & \begin{CJK}{UTF8}{gbsn} 牛奶可以和咖啡一起喝。\end{CJK} \\
              &(Milk can be drunk with coffee.) \\

    \midrule
    Normal prompt & \emph{You are a helpful assistant.} \\ 
    Response & \begin{CJK}{UTF8}{gbsn} 牛奶没有了，去喝咖啡吧。\end{CJK} \\
              &(There's no milk, let's drink coffee.) \\
    \midrule
    + Task prompt & \emph{\textbf{Use your expertise to provide the most precise translation.}} \\ 
    Response & \begin{CJK}{UTF8}{gbsn} 牛奶和咖啡都可以喝。\end{CJK} \\
              &(You can drink both milk and coffee.) \\
    \midrule
    + Format prompt & \emph{\textbf{Answer with one single sentence.}}  \\ 
    Response & \begin{CJK}{UTF8}{gbsn} 牛奶和咖啡可以一起喝。\end{CJK} \\
              &(Milk and coffee can be drunk together.) \\
    \bottomrule
  \end{tabular}
}
\caption{Qualitative evaluation on \textbf{prompting strategies}. As progressively more specific prompts are incorporated, the accuracy of the translation result significantly improves.}  
\label{subtab:format_prompts_diff} 
\end{table} 
\section{Conclusion}
\label{sec:discussion}
In this paper, we introduce LLaVA-SLT, a pioneering framework for sign language translation that leverages the paradigm of Large Multimodal Models. Our approach integrates linguistic continued pretraining, visual contrastive pretraining, and visual language tuning to create a robust and scalable solution for gloss-free SLT.
The linguistic continued pretraining stage adapts general-purpose LLMs to the sign language domain, enhancing their textual linguistic knowledge. The visual contrastive pretraining aligns the visual encoder with a large-scale pretrained text encoder, learning robust visual language representations. Finally, the visual language tuning efficiently maps these visual embeddings into the LLM token embedding space, enabling effective downstream SLT task.
Our extensive experiments demonstrate that LLaVA-SLT achieves state-of-the-art performance in gloss-free SLT, significantly narrowing the gap with gloss-based methods. We believe that LLaVA-SLT represents a significant advancement in the field, with the potential to greatly enhance communication between hard-of-hearing and hearing people.

\clearpage


{
    \small
    \bibliographystyle{ieeenat_fullname}
    \bibliography{main}

\begin{thebibliography}{123}
\providecommand{\natexlab}[1]{#1}
\providecommand{\url}[1]{\texttt{#1}}
\expandafter\ifx\csname urlstyle\endcsname\relax
  \providecommand{\doi}[1]{doi: #1}\else
  \providecommand{\doi}{doi: \begingroup \urlstyle{rm}\Url}\fi

\bibitem[Albanie et~al.(2020)Albanie, Varol, Momeni, et~al.]{albanie2020bsl-1k}
S Albanie, G Varol, L Momeni, et~al.
\newblock Bsl-1k: Scaling up co-articulated sign language recognition using mouthing cues.
\newblock In \emph{Computer Vision–ECCV 2020: 16th European Conference, Glasgow, UK, August 23–28, 2020, Proceedings, Part XI}, pages 35--53. Springer International Publishing, 2020.

\bibitem[Bai et~al.(2023)Bai, Bai, Chu, Cui, Dang, Deng, Fan, Ge, Han, Huang, et~al.]{bai2023qwen}
Jinze Bai, Shuai Bai, Yunfei Chu, Zeyu Cui, Kai Dang, Xiaodong Deng, Yang Fan, Wenbin Ge, Yu Han, Fei Huang, et~al.
\newblock Qwen technical report.
\newblock \emph{arXiv preprint arXiv:2309.16609}, 2023.

\bibitem[\begin{CJK}{UTF8}{gbsn}Li et~al.(2023)\begin{CJK}{UTF8}{gbsn}Li, Luo, Li, and Xi]{LiWenyu2023}
Wenyu \begin{CJK}{UTF8}{gbsn}Li, Zhizeng Luo, Wenguo Li, and Xugang\end{CJK} Xi.
\newblock \begin{CJK}{UTF8}{gbsn}chinese sign language recognition based on surface electromyography and motion information\end{CJK}.
\newblock \emph{\begin{CJK}{UTF8}{gbsn}PLOS ONE\end{CJK}}, 18\penalty0 (12):\penalty0 1--15, 2023.

\bibitem[\begin{CJK}{UTF8}{gbsn}中国聋人协会\end{CJK}(2003)]{ZhongguoShouyu}
\begin{CJK}{UTF8}{gbsn}中国聋人协会\end{CJK}.
\newblock \emph{\begin{CJK}{UTF8}{gbsn}中国手语（上下修订版）\end{CJK}}.
\newblock \begin{CJK}{UTF8}{gbsn}华夏出版社\end{CJK}, 2003.

\bibitem[\begin{CJK}{UTF8}{gbsn}中国聋人协会\end{CJK}(2019)]{GuoTongyongShouyu}
\begin{CJK}{UTF8}{gbsn}中国聋人协会\end{CJK}.
\newblock \emph{\begin{CJK}{UTF8}{gbsn}国家通用手语词典\end{CJK}}.
\newblock \begin{CJK}{UTF8}{gbsn}华夏出版社\end{CJK}, 2019.

\bibitem[\begin{CJK}{UTF8}{gbsn}倪兰\end{CJK}(2015)]{NiLan2015}
\begin{CJK}{UTF8}{gbsn}倪兰\end{CJK}.
\newblock \emph{\begin{CJK}{UTF8}{gbsn}中国手语动词研究\end{CJK}}.
\newblock 2015.

\bibitem[\begin{CJK}{UTF8}{gbsn}刘学达\end{CJK}(2022)]{LiuXueda2022}
\begin{CJK}{UTF8}{gbsn}刘学达\end{CJK}.
\newblock \begin{CJK}{UTF8}{gbsn}中国手语语料库高频词初步分析及标注探讨\end{CJK}.
\newblock Master's thesis, \begin{CJK}{UTF8}{gbsn}上海外国语大学\end{CJK}, 2022.

\bibitem[\begin{CJK}{UTF8}{gbsn}吕会华\end{CJK}(2017)]{LvHuaihua2017}
\begin{CJK}{UTF8}{gbsn}吕会华\end{CJK}.
\newblock \begin{CJK}{UTF8}{gbsn}中国手语中的“指点”手势研究\end{CJK}.
\newblock \emph{\begin{CJK}{UTF8}{gbsn}绥化学院学报\end{CJK}}, 37\penalty0 (07):\penalty0 8--13, 2017.

\bibitem[\begin{CJK}{UTF8}{gbsn}张吉生\end{CJK}(2019)]{ZhangJisheng2019}
\begin{CJK}{UTF8}{gbsn}张吉生\end{CJK}.
\newblock \emph{\begin{CJK}{UTF8}{gbsn}上海手语音系\end{CJK}}.
\newblock 2019.

\bibitem[\begin{CJK}{UTF8}{gbsn}林皓\end{CJK}(2018)]{LinHao2018}
\begin{CJK}{UTF8}{gbsn}林皓\end{CJK}.
\newblock \begin{CJK}{UTF8}{gbsn}中国手语一般疑问句中疑问手控标记研究\end{CJK}.
\newblock \emph{\begin{CJK}{UTF8}{gbsn}语言研究集刊\end{CJK}}, \penalty0 (01):\penalty0 241--255+379, 2018.

\bibitem[\begin{CJK}{UTF8}{gbsn}潘一\end{CJK}(2005)]{ShouyuHuihua}
\begin{CJK}{UTF8}{gbsn}潘一\end{CJK}.
\newblock \emph{\begin{CJK}{UTF8}{gbsn}手语会话\end{CJK}}.
\newblock \begin{CJK}{UTF8}{gbsn}高等教育出版社\end{CJK}, 2005.

\bibitem[\begin{CJK}{UTF8}{gbsn}王晓霞\end{CJK}(2020)]{WangXiaoxia2020}
\begin{CJK}{UTF8}{gbsn}王晓霞\end{CJK}.
\newblock \begin{CJK}{UTF8}{gbsn}中国手语音系学中的音节结构研究\end{CJK}.
\newblock Master's thesis, \begin{CJK}{UTF8}{gbsn}西安外国语大学\end{CJK}, 2020.

\bibitem[\begin{CJK}{UTF8}{gbsn}郑璇\end{CJK}(2015)]{ShouyuJichu}
\begin{CJK}{UTF8}{gbsn}郑璇\end{CJK}.
\newblock \emph{\begin{CJK}{UTF8}{gbsn}手语基础教程\end{CJK}}.
\newblock \begin{CJK}{UTF8}{gbsn}华东师范大学出版社\end{CJK}, 2015.

\bibitem[Biderman et~al.(2024)Biderman, Portes, Ortiz, Paul, Greengard, Jennings, King, Havens, Chiley, Frankle, et~al.]{biderman2024lora}
Dan Biderman, Jacob Portes, Jose Javier~Gonzalez Ortiz, Mansheej Paul, Philip Greengard, Connor Jennings, Daniel King, Sam Havens, Vitaliy Chiley, Jonathan Frankle, et~al.
\newblock Lora learns less and forgets less.
\newblock \emph{arXiv preprint arXiv:2405.09673}, 2024.

\bibitem[Blog(2024)]{deaflepuff_chinese_sign_language}
Deaflepuff Blog.
\newblock Making themselves heard: Chinese sign language \& deaf china online.
\newblock \url{https://deaflepuff.tumblr.com/post/142025989481/making-themselves-heard-chinese-sign-language}, 2024.
\newblock Accessed: November 20, 2024.

\bibitem[Blogs(2024)]{csl_sign_language_blogs}
Sign~Language Blogs.
\newblock Csl, the revolutionary chinese sign language.
\newblock \url{https://signlanguage.blog/chinese-sign-language-csl/}, 2024.
\newblock Accessed: November 20, 2024.

\bibitem[Camgoz et~al.(2018)Camgoz, Hadfield, Koller, et~al.]{camgoz2018neural}
N~C Camgoz, S Hadfield, O Koller, et~al.
\newblock Neural sign language translation.
\newblock In \emph{Proceedings of the IEEE Conference on Computer Vision and Pattern Recognition}, pages 7784--7793, 2018.

\bibitem[Camgoz et~al.(2020)Camgoz, Koller, Hadfield, et~al.]{camgoz2020sign}
N~C Camgoz, O Koller, S Hadfield, et~al.
\newblock Sign language transformers: Joint end-to-end sign language recognition and translation.
\newblock In \emph{Proceedings of the IEEE/CVF Conference on Computer Vision and Pattern Recognition}, pages 10023--10033, 2020.

\bibitem[CGTN(2019{\natexlab{a}})]{idsl_cgtn_dreams_rights}
CGTN.
\newblock Idsl: Unleashing dreams and protecting rights of the hearing-impaired.
\newblock \url{https://news.cgtn.com/news/2019-09-23/International-Day-of-Sign-Languages-Unleashing-dreams-and-protecting-rights-of-the-hearing-impaired-K6HMsDJxfi/index.html}, 2019{\natexlab{a}}.
\newblock Accessed: November 20, 2024.

\bibitem[CGTN(2019{\natexlab{b}})]{idsl_cgtn_history}
CGTN.
\newblock Idsl: China's sign language history dates back to tang dynasty period.
\newblock \url{https://news.cgtn.com/news/2019-09-23/IDSL-China-s-sign-language-history-dates-back-to-Tang-Dynasty-period-KamfGp0Q1y/index.html}, 2019{\natexlab{b}}.
\newblock Accessed: November 20, 2024.

\bibitem[Chen et~al.(2024{\natexlab{a}})Chen, Wang, Guo, et~al.]{chen2024signvtcl}
H Chen, J Wang, Z Guo, et~al.
\newblock Signvtcl: Multi-modal continuous sign language recognition enhanced by visual-textual contrastive learning.
\newblock \emph{arXiv preprint arXiv:2401.11847}, 2024{\natexlab{a}}.

\bibitem[Chen et~al.(2024{\natexlab{b}})Chen, Chen, Zhang, Li, Yu, Fei, Zhu, Fan, and Chen]{chen2024ll3da}
Sijin Chen, Xin Chen, Chi Zhang, Mingsheng Li, Gang Yu, Hao Fei, Hongyuan Zhu, Jiayuan Fan, and Tao Chen.
\newblock Ll3da: Visual interactive instruction tuning for omni-3d understanding reasoning and planning.
\newblock In \emph{Proceedings of the IEEE/CVF Conference on Computer Vision and Pattern Recognition}, pages 26428--26438, 2024{\natexlab{b}}.

\bibitem[Chen et~al.(2022{\natexlab{a}})Chen, Wei, Sun, et~al.]{chen2022simple}
Y Chen, F Wei, X Sun, et~al.
\newblock A simple multi-modality transfer learning baseline for sign language translation.
\newblock In \emph{Proceedings of the IEEE/CVF Conference on Computer Vision and Pattern Recognition}, pages 5120--5130, 2022{\natexlab{a}}.

\bibitem[Chen et~al.(2022{\natexlab{b}})Chen, Zuo, Wei, et~al.]{chen2022two-stream}
Y Chen, R Zuo, F Wei, et~al.
\newblock Two-stream network for sign language recognition and translation.
\newblock \emph{Advances in Neural Information Processing Systems}, 35:\penalty0 17043--17056, 2022{\natexlab{b}}.

\bibitem[Chen et~al.(2024{\natexlab{c}})Chen, Zhou, Li, Wan, Lei, Jiang, Lu, and Zhao]{chen2024factorized}
Zhigang Chen, Benjia Zhou, Jun Li, Jun Wan, Zhen Lei, Ning Jiang, Quan Lu, and Guoqing Zhao.
\newblock Factorized learning assisted with large language model for gloss-free sign language translation.
\newblock In \emph{Proceedings of the 2024 Joint International Conference on Computational Linguistics, Language Resources and Evaluation (LREC-COLING 2024)}, pages 7071--7081, 2024{\natexlab{c}}.

\bibitem[Cheng et~al.(2020)Cheng, Yang, Chen, et~al.]{cheng2020fully}
K~L Cheng, Z Yang, Q Chen, et~al.
\newblock Fully convolutional networks for continuous sign language recognition.
\newblock In \emph{Computer Vision–ECCV 2020: 16th European Conference, Glasgow, UK, August 23–28, 2020, Proceedings, Part XXIV}, pages 697--714. Springer International Publishing, 2020.

\bibitem[{\begin{CJK}{UTF8}{gbsn}吕会华, 王红英\end{CJK}}(2017)]{LvHuaihuaWangHongying2017}
{\begin{CJK}{UTF8}{gbsn}吕会华, 王红英\end{CJK}}.
\newblock \begin{CJK}{UTF8}{gbsn}中国手语定中短语的语序\end{CJK}.
\newblock \emph{\begin{CJK}{UTF8}{gbsn}中国听力语言康复科学杂志\end{CJK}}, 15\penalty0 (06):\penalty0 459--463, 2017.

\bibitem[{\begin{CJK}{UTF8}{gbsn}姚登峰, 江铭虎, 张荣兴, 阿布都克力木·阿布力孜\end{CJK}}(2018)]{YaoDengfeng2018}
{\begin{CJK}{UTF8}{gbsn}姚登峰, 江铭虎, 张荣兴, 阿布都克力木·阿布力孜\end{CJK}}.
\newblock \begin{CJK}{UTF8}{gbsn}论中国手语的分类词谓语\end{CJK}.
\newblock \emph{\begin{CJK}{UTF8}{gbsn}中文信息学报\end{CJK}}, 32\penalty0 (03):\penalty0 1--8, 2018.

\bibitem[{\begin{CJK}{UTF8}{gbsn}张艳琼, 朱兆松, 张胜伟, 赵晓驰\end{CJK}}(2024)]{ZhangYanqun2024}
{\begin{CJK}{UTF8}{gbsn}张艳琼, 朱兆松, 张胜伟, 赵晓驰\end{CJK}}.
\newblock \begin{CJK}{UTF8}{gbsn}中国手语基本手势语义知识组织研究\end{CJK}.
\newblock \emph{\begin{CJK}{UTF8}{gbsn}情报科学\end{CJK}}, 42\penalty0 (03):\penalty0 118--128, 2024.

\bibitem[{\begin{CJK}{UTF8}{gbsn}文旭, 曹阳\end{CJK}}(2022)]{WenXu2022}
{\begin{CJK}{UTF8}{gbsn}文旭, 曹阳\end{CJK}}.
\newblock \begin{CJK}{UTF8}{gbsn}中国手语相同音系参数下的词义拓展和相关性构建\end{CJK}.
\newblock \emph{\begin{CJK}{UTF8}{gbsn}西北工业大学学报(社会科学版)\end{CJK}}, \penalty0 (01):\penalty0 77--84, 2022.

\bibitem[Community(2024)]{ni_lan_interview_2024}
Tencent Cloud~Developer Community.
\newblock \begin{CJK}{UTF8}{gbsn} 专访上海大学倪兰教授：语言学与手语识别技术的融合突破，解锁交流障碍 - gair live\end{CJK}.
\newblock \url{https://cloud.tencent.com/developer/article/2397557}, 2024.
\newblock Accessed: November 20, 2024.

\bibitem[Contributors(2024{\natexlab{a}})]{sign_language_baidu}
Baidu~Baike Contributors.
\newblock \begin{CJK}{UTF8}{gbsn} 手语（聋人交流方式-百度百科\end{CJK}.
\newblock \url{https://baike.baidu.com/item/\%E6\%89\%8B\%E8\%AF\%AD/67906?fr=ge_ala}, 2024{\natexlab{a}}.
\newblock Accessed: November 20, 2024.

\bibitem[Contributors(2024{\natexlab{b}})]{asl_vs_csl_quora}
Quora Contributors.
\newblock Is american sign language different from chinese sign language?
\newblock \url{https://www.quora.com/Is-American-Sign-Language-different-from-Chinese-Sign-Language}, 2024{\natexlab{b}}.
\newblock Accessed: November 20, 2024.

\bibitem[Contributors(2024{\natexlab{c}})]{chinese_sign_language_wikipedia}
Wikipedia Contributors.
\newblock \begin{CJK}{UTF8}{gbsn} 中国手语 - 维基百科，自由的百科全书\end{CJK}.
\newblock \url{https://zh.wikipedia.org/zh-hans/\%E4\%B8\%AD\%E5\%9C\%8B\%E6\%89\%8B\%E8\%AA\%9E}, 2024{\natexlab{c}}.
\newblock Accessed: November 20, 2024.

\bibitem[Contributors(2024{\natexlab{d}})]{csl_wikipedia_english}
Wikipedia Contributors.
\newblock Chinese sign language - wikipedia.
\newblock \url{https://en.wikipedia.org/wiki/Chinese_Sign_Language}, 2024{\natexlab{d}}.
\newblock Accessed: November 20, 2024.

\bibitem[Contributors(2024{\natexlab{e}})]{sign_language_wikipedia}
Wikipedia Contributors.
\newblock \begin{CJK}{UTF8}{gbsn} 手语 - 维基百科，自由的百科全书\end{CJK}.
\newblock \url{https://zh.wikipedia.org/wiki/\%E6\%89\%8B\%E8\%AA\%9E}, 2024{\natexlab{e}}.
\newblock Accessed: November 20, 2024.

\bibitem[Cui et~al.(2019)Cui, Liu, and Zhang]{cui2019deep}
R Cui, H Liu, and C Zhang.
\newblock A deep neural framework for continuous sign language recognition by iterative training.
\newblock \emph{IEEE Transactions on Multimedia}, 21\penalty0 (7):\penalty0 1880--1891, 2019.

\bibitem[Desai et~al.(2024)Desai, Berger, Minakov, et~al.]{desai2024asl-citizen}
A Desai, L Berger, F Minakov, et~al.
\newblock Asl citizen: A community-sourced dataset for advancing isolated sign language recognition.
\newblock \emph{Advances in Neural Information Processing Systems}, 36, 2024.

\bibitem[Duarte et~al.(2021)Duarte, Palaskar, Ventura, et~al.]{duarte2021how2sign}
A Duarte, S Palaskar, L Ventura, et~al.
\newblock How2sign: A large-scale multimodal dataset for continuous american sign language.
\newblock In \emph{Proceedings of the IEEE/CVF Conference on Computer Vision and Pattern Recognition}, pages 2735--2744, 2021.

\bibitem[Dubey et~al.(2024)Dubey, Jauhri, Pandey, Kadian, Al-Dahle, Letman, Mathur, Schelten, Yang, Fan, et~al.]{dubey2024llama}
Abhimanyu Dubey, Abhinav Jauhri, Abhinav Pandey, Abhishek Kadian, Ahmad Al-Dahle, Aiesha Letman, Akhil Mathur, Alan Schelten, Amy Yang, Angela Fan, et~al.
\newblock The llama 3 herd of models.
\newblock \emph{arXiv preprint arXiv:2407.21783}, 2024.

\bibitem[Feng et~al.(2024)Feng, Lin, Dwivedi, et~al.]{feng2024chatpose}
Y Feng, J Lin, S~K Dwivedi, et~al.
\newblock Chatpose: Chatting about 3d human pose.
\newblock In \emph{Proceedings of the IEEE/CVF Conference on Computer Vision and Pattern Recognition}, pages 2093--2103, 2024.

\bibitem[Forster et~al.(2014)Forster, Schmidt, Koller, et~al.]{forster2014extensions}
J Forster, C Schmidt, O Koller, et~al.
\newblock Extensions of the sign language recognition and translation corpus rwth-phoenix-weather.
\newblock In \emph{LREC}, pages 1911--1916, 2014.

\bibitem[Forums(2024)]{chinese_sign_language_forums}
Chinese Forums.
\newblock \begin{CJK}{UTF8}{gbsn} 中国手语/chinese sign language - non-mandarin chinese - chinese-forums\end{CJK}.
\newblock \url{https://www.chinese-forums.com/forums/topic/50579-\%E4\%B8\%AD\%E5\%9B\%BD\%E6\%89\%8B\%E8\%AF\%ADchinese-sign-language/}, 2024.
\newblock Accessed: November 20, 2024.

\bibitem[Gao et~al.(2004)Gao, Fang, Zhao, et~al.]{gao2004chinese}
W Gao, G Fang, D Zhao, et~al.
\newblock A chinese sign language recognition system based on sofm/srn/hmm.
\newblock \emph{Pattern Recognition}, 37\penalty0 (12):\penalty0 2389--2402, 2004.

\bibitem[Gong et~al.(2024)Gong, Foo, He, et~al.]{gong2024llms}
J Gong, L~G Foo, Y He, et~al.
\newblock Llms are good sign language translators.
\newblock In \emph{Proceedings of the IEEE/CVF Conference on Computer Vision and Pattern Recognition}, pages 18362--18372, 2024.

\bibitem[Han et~al.(2024)Han, Gong, Zhang, Wang, Zhang, Lin, Qiao, Gao, and Yue]{han2024onellm}
Jiaming Han, Kaixiong Gong, Yiyuan Zhang, Jiaqi Wang, Kaipeng Zhang, Dahua Lin, Yu Qiao, Peng Gao, and Xiangyu Yue.
\newblock Onellm: One framework to align all modalities with language.
\newblock In \emph{Proceedings of the IEEE/CVF Conference on Computer Vision and Pattern Recognition}, pages 26584--26595, 2024.

\bibitem[He et~al.(2024)He, Li, Jang, et~al.]{he2024ma-lmm}
B He, H Li, Y~K Jang, et~al.
\newblock Ma-lmm: Memory-augmented large multimodal model for long-term video understanding.
\newblock In \emph{Proceedings of the IEEE/CVF Conference on Computer Vision and Pattern Recognition}, pages 13504--13514, 2024.

\bibitem[Hu et~al.(202)Hu, Wallis, Allen-Zhu, Li, Wang, Wang, Chen, et~al.]{hu2021lora}
Edward~J Hu, Phillip Wallis, Zeyuan Allen-Zhu, Yuanzhi Li, Shean Wang, Lu Wang, Weizhu Chen, et~al.
\newblock Lora: Low-rank adaptation of large language models.
\newblock In \emph{International Conference on Learning Representations}, 202.

\bibitem[Hu et~al.(2021)Hu, Zhao, Zhou, et~al.]{hu2021signbert}
H Hu, W Zhao, W Zhou, et~al.
\newblock Signbert: Pre-training of hand-model-aware representation for sign language recognition.
\newblock In \emph{Proceedings of the IEEE/CVF International Conference on Computer Vision}, pages 11087--11096, 2021.

\bibitem[Hu et~al.(2023)Hu, Gao, Liu, et~al.]{hu2023self-emphasizing}
L Hu, L Gao, Z Liu, et~al.
\newblock Self-emphasizing network for continuous sign language recognition.
\newblock In \emph{Proceedings of the AAAI Conference on Artificial Intelligence}, pages 854--862, 2023.

\bibitem[Hu et~al.(2023a)Hu, Gao, Liu, et~al.]{hu2023continuous}
L Hu, L Gao, Z Liu, et~al.
\newblock Continuous sign language recognition with correlation network.
\newblock In \emph{Proceedings of the IEEE/CVF Conference on Computer Vision and Pattern Recognition}, pages 2529--2539, 2023a.

\bibitem[Huang et~al.(2018)Huang, Zhou, Zhang, et~al.]{huang2018video-based}
J Huang, W Zhou, Q Zhang, et~al.
\newblock Video-based sign language recognition without temporal segmentation.
\newblock In \emph{Proceedings of the AAAI Conference on Artificial Intelligence}, 2018.

\bibitem[Huang et~al.(2024)Huang, Li, Yang, et~al.]{huang2024audiogpt}
R Huang, M Li, D Yang, et~al.
\newblock Audiogpt: Understanding and generating speech, music, sound, and talking head.
\newblock In \emph{Proceedings of the AAAI Conference on Artificial Intelligence}, pages 23802--23804, 2024.

\bibitem[Hui et~al.(2024)Hui, Yang, Cui, Yang, Liu, Zhang, Liu, Zhang, Yu, Dang, et~al.]{hui2024qwen2}
Binyuan Hui, Jian Yang, Zeyu Cui, Jiaxi Yang, Dayiheng Liu, Lei Zhang, Tianyu Liu, Jiajun Zhang, Bowen Yu, Kai Dang, et~al.
\newblock Qwen2. 5-coder technical report.
\newblock \emph{arXiv preprint arXiv:2409.12186}, 2024.

\bibitem[Imashev et~al.(2020)Imashev, Mukushev, Kimmelman, et~al.]{imashev2020dataset}
A Imashev, M Mukushev, V Kimmelman, et~al.
\newblock A dataset for linguistic understanding, visual evaluation, and recognition of sign languages: The k-rsl.
\newblock In \emph{Proceedings of the 24th Conference on Computational Natural Language Learning}, pages 631--640, 2020.

\bibitem[Imran et~al.(2024)Imran, Khan, Biswas, and Islam]{imran2024llasa}
Sheikh~Asif Imran, Mohammad Nur~Hossain Khan, Subrata Biswas, and Bashima Islam.
\newblock Llasa: Large multimodal agent for human activity analysis through wearable sensors.
\newblock \emph{arXiv preprint arXiv:2406.14498}, 2024.

\bibitem[Jiang et~al.(2024)Jiang, Sant, Moryossef, M{\"u}ller, Sennrich, and Ebling]{jiang2024signclip}
Zifan Jiang, Gerard Sant, Amit Moryossef, Mathias M{\"u}ller, Rico Sennrich, and Sarah Ebling.
\newblock Signclip: Connecting text and sign language by contrastive learning.
\newblock In \emph{Proceedings of the 2024 Conference on Empirical Methods in Natural Language Processing}, pages 9171--9193, 2024.

\bibitem[Joze and Koller(2018)]{joze2018ms-asl}
H~R~V Joze and O Koller.
\newblock Ms-asl: A large-scale dataset and benchmark for understanding american sign language.
\newblock \emph{arXiv preprint arXiv:1812.01053}, 2018.

\bibitem[Ko et~al.(2019)Ko, Kim, Jung, et~al.]{ko2019neural}
S~K Ko, C~J Kim, H Jung, et~al.
\newblock Neural sign language translation based on human keypoint estimation.
\newblock \emph{Applied Sciences}, 9\penalty0 (13):\penalty0 2683, 2019.

\bibitem[Koller et~al.(2016)Koller, Zargaran, Ney, et~al.]{koller2016deep}
O Koller, S Zargaran, H Ney, et~al.
\newblock Deep sign: Hybrid cnn-hmm for continuous sign language recognition.
\newblock In \emph{BMVC}, pages 136.1--136.12, 2016.

\bibitem[Koller et~al.(2017a)Koller, Zargaran, and Ney]{koller2017re-sign}
O Koller, S Zargaran, and H Ney.
\newblock Re-sign: Re-aligned end-to-end sequence modelling with deep recurrent cnn-hmms.
\newblock In \emph{Proceedings of the IEEE Conference on Computer Vision and Pattern Recognition}, pages 4297--4305, 2017a.

\bibitem[Li et~al.(2020)Li, Yu, Xu, et~al.]{li2020transferring}
D Li, X Yu, C Xu, et~al.
\newblock Transferring cross-domain knowledge for video sign language recognition.
\newblock In \emph{Proceedings of the IEEE/CVF Conference on Computer Vision and Pattern Recognition}, pages 6205--6214, 2020.

\bibitem[Li et~al.(2020a)Li, Rodriguez, Yu, et~al.]{li2020word-level}
D Li, C Rodriguez, X Yu, et~al.
\newblock Word-level deep sign language recognition from video: A new large-scale dataset and methods comparison.
\newblock In \emph{Proceedings of the IEEE/CVF Winter Conference on Applications of Computer Vision}, pages 1459--1469, 2020a.

\bibitem[Li et~al.(2020b)Li, Xu, Yu, et~al.]{li2020tspnet}
D Li, C Xu, X Yu, et~al.
\newblock Tspnet: Hierarchical feature learning via temporal semantic pyramid for sign language translation.
\newblock In \emph{Advances in Neural Information Processing Systems}, pages 12034--12045, 2020b.

\bibitem[Li et~al.(2024{\natexlab{a}})Li, Zhang, Zhang, Zhang, Li, Li, Ma, and Li]{li2024llava}
Feng Li, Renrui Zhang, Hao Zhang, Yuanhan Zhang, Bo Li, Wei Li, Zejun Ma, and Chunyuan Li.
\newblock Llava-next-interleave: Tackling multi-image, video, and 3d in large multimodal models.
\newblock \emph{arXiv preprint arXiv:2407.07895}, 2024{\natexlab{a}}.

\bibitem[Li et~al.(2024{\natexlab{b}})Li, Zhang, Zhang, et~al.]{li2024llava-next-interleave}
F Li, R Zhang, H Zhang, et~al.
\newblock Llava-next-interleave: Tackling multi-image, video, and 3d in large multimodal models.
\newblock \emph{arXiv preprint arXiv:2407.07895}, 2024{\natexlab{b}}.

\bibitem[Li et~al.(2023)Li, He, Wang, et~al.]{li2023videochat}
K~C Li, Y He, Y Wang, et~al.
\newblock Videochat: Chat-centric video understanding.
\newblock \emph{arXiv preprint arXiv:2305.06355}, 2023.

\bibitem[Liang et~al.(2024)Liang, Bao, Zhang, Ren, Xu, Yang, Chen, Yu, and Xu]{liang2024omg}
Han Liang, Jiacheng Bao, Ruichi Zhang, Sihan Ren, Yuecheng Xu, Sibei Yang, Xin Chen, Jingyi Yu, and Lan Xu.
\newblock Omg: Towards open-vocabulary motion generation via mixture of controllers.
\newblock In \emph{Proceedings of the IEEE/CVF Conference on Computer Vision and Pattern Recognition}, pages 482--493, 2024.

\bibitem[Lin et~al.(2023{\natexlab{a}})Lin, Ye, Zhu, Cui, Ning, Jin, and Yuan]{lin2023video}
Bin Lin, Yang Ye, Bin Zhu, Jiaxi Cui, Munan Ning, Peng Jin, and Li Yuan.
\newblock Video-llava: Learning united visual representation by alignment before projection.
\newblock \emph{arXiv preprint arXiv:2311.10122}, 2023{\natexlab{a}}.

\bibitem[Lin(2004)]{lin2004rouge}
Chin-Yew Lin.
\newblock Rouge: A package for automatic evaluation of summaries.
\newblock In \emph{Text summarization branches out}, pages 74--81, 2004.

\bibitem[Lin et~al.(2024)Lin, Feng, Liu, et~al.]{lin2024chathuman}
J Lin, Y Feng, W Liu, et~al.
\newblock Chathuman: Language-driven 3d human understanding with retrieval-augmented tool reasoning.
\newblock \emph{arXiv preprint arXiv:2405.04533}, 2024.

\bibitem[Lin et~al.(2023{\natexlab{b}})Lin, Wang, Zhu, Sun, Zhang, and Yang]{lin2023gloss}
Kezhou Lin, Xiaohan Wang, Linchao Zhu, Ke Sun, Bang Zhang, and Yi Yang.
\newblock Gloss-free end-to-end sign language translation.
\newblock \emph{arXiv preprint arXiv:2305.12876}, 2023{\natexlab{b}}.

\bibitem[Liu et~al.(2024{\natexlab{a}})Liu, Li, Li, and Lee]{liu2024improved}
Haotian Liu, Chunyuan Li, Yuheng Li, and Yong~Jae Lee.
\newblock Improved baselines with visual instruction tuning.
\newblock In \emph{Proceedings of the IEEE/CVF Conference on Computer Vision and Pattern Recognition}, pages 26296--26306, 2024{\natexlab{a}}.

\bibitem[Liu et~al.(2024{\natexlab{b}})Liu, Li, Wu, et~al.]{liu2024visual}
H Liu, C Li, Q Wu, et~al.
\newblock Visual instruction tuning.
\newblock In \emph{Advances in Neural Information Processing Systems}, 2024{\natexlab{b}}.

\bibitem[Liu et~al.(2020)Liu, Gu, Goyal, Li, Edunov, Ghazvininejad, Lewis, and Zettlemoyer]{liu2020multilingual}
Yinhan Liu, Jiatao Gu, Naman Goyal, Xian Li, Sergey Edunov, Marjan Ghazvininejad, Mike Lewis, and Luke Zettlemoyer.
\newblock Multilingual denoising pre-training for neural machine translation.
\newblock \emph{Transactions of the Association for Computational Linguistics}, 2020.

\bibitem[Loshchilov(2017)]{loshchilov2017decoupled}
I Loshchilov.
\newblock Decoupled weight decay regularization.
\newblock \emph{arXiv preprint arXiv:1711.05101}, 2017.

\bibitem[Lyu et~al.(2023)Lyu, Wu, Wang, Huang, Liu, Du, Shi, and Tu]{lyu2023macaw}
Chenyang Lyu, Minghao Wu, Longyue Wang, Xinting Huang, Bingshuai Liu, Zefeng Du, Shuming Shi, and Zhaopeng Tu.
\newblock Macaw-llm: Multi-modal language modeling with image, audio, video, and text integration.
\newblock \emph{arXiv preprint arXiv:2306.09093}, 2023.

\bibitem[Maaz et~al.(2023)Maaz, Rasheed, Khan, et~al.]{maaz2023video-chatgpt}
M Maaz, H Rasheed, S Khan, et~al.
\newblock Video-chatgpt: Towards detailed video understanding via large vision and language models.
\newblock \emph{arXiv preprint arXiv:2306.05424}, 2023.

\bibitem[Min et~al.(2021)Min, Hao, Chai, et~al.]{min2021visual}
Y Min, A Hao, X Chai, et~al.
\newblock Visual alignment constraint for continuous sign language recognition.
\newblock In \emph{Proceedings of the IEEE/CVF International Conference on Computer Vision}, pages 11542--11551, 2021.

\bibitem[Online(2024)]{sign_language_learning_bmcx}
Sign~Language Online.
\newblock \begin{CJK}{UTF8}{gbsn} 手语翻译 - 学习手语 - 手语图解 - 手语查询\end{CJK}.
\newblock \url{https://shouyu.bmcx.com/}, 2024.
\newblock Accessed: November 20, 2024.

\bibitem[Oquab et~al.(2023)Oquab, Darcet, Moutakanni, Vo, Szafraniec, Khalidov, Fernandez, HAZIZA, Massa, El-Nouby, et~al.]{oquab2023dinov2}
Maxime Oquab, Timoth{\'e}e Darcet, Th{\'e}o Moutakanni, Huy~V Vo, Marc Szafraniec, Vasil Khalidov, Pierre Fernandez, Daniel HAZIZA, Francisco Massa, Alaaeldin El-Nouby, et~al.
\newblock Dinov2: Learning robust visual features without supervision.
\newblock \emph{Transactions on Machine Learning Research}, 2023.

\bibitem[Papadimitriou and Potamianos(2020)]{papadimitriou2020multimodal}
K Papadimitriou and G Potamianos.
\newblock Multimodal sign language recognition via temporal deformable convolutional sequence learning.
\newblock In \emph{Interspeech}, pages 2752--2756, 2020.

\bibitem[Papineni et~al.(2002)Papineni, Roukos, Ward, and Zhu]{papineni2002bleu}
Kishore Papineni, Salim Roukos, Todd Ward, and Wei-Jing Zhu.
\newblock Bleu: a method for automatic evaluation of machine translation.
\newblock In \emph{Proceedings of the 40th annual meeting of the Association for Computational Linguistics}, pages 311--318, 2002.

\bibitem[Radford et~al.(2021)Radford, Kim, Hallacy, Ramesh, Goh, Agarwal, Sastry, Askell, Mishkin, Clark, et~al.]{radford2021learning}
Alec Radford, Jong~Wook Kim, Chris Hallacy, Aditya Ramesh, Gabriel Goh, Sandhini Agarwal, Girish Sastry, Amanda Askell, Pamela Mishkin, Jack Clark, et~al.
\newblock Learning transferable visual models from natural language supervision.
\newblock In \emph{International conference on machine learning}, pages 8748--8763. PMLR, 2021.

\bibitem[Rajbhandari et~al.(2020)Rajbhandari, Rasley, Ruwase, and He]{rajbhandari2020zero}
Samyam Rajbhandari, Jeff Rasley, Olatunji Ruwase, and Yuxiong He.
\newblock Zero: Memory optimizations toward training trillion parameter models.
\newblock In \emph{SC20: International Conference for High Performance Computing, Networking, Storage and Analysis}, pages 1--16. IEEE, 2020.

\bibitem[Ren et~al.(2024)Ren, Yao, Yang, and Kang]{RenTianyu2024}
Tianyu Ren, Dengfeng Yao, Chaoran Yang, and Xinchen Kang.
\newblock The influence of chinese characters on chinese sign language.
\newblock 23\penalty0 (1), 2024.

\bibitem[School(2024)]{csl_french_ltl_school}
LTL~Language School.
\newblock Connaissez-vous la langue des signes chinoise? un guide complet sur \begin{CJK}{UTF8}{gbsn} 中国手语\end{CJK}.
\newblock \url{https://ltl-school.fr/langue-des-signes-chinoise/}, 2024.
\newblock Accessed: November 20, 2024.

\bibitem[Sincan and Keles(2020)]{sincand2020autsl}
O~M Sincan and H~Y Keles.
\newblock Autsl: A large scale multi-modal turkish sign language dataset and baseline methods.
\newblock \emph{IEEE Access}, 8:\penalty0 181340--181355, 2020.

\bibitem[Sridhar et~al.(2020)Sridhar, Ganesan, Kumar, et~al.]{sridhar2020include}
A Sridhar, R~G Ganesan, P Kumar, et~al.
\newblock Include: A large scale dataset for indian sign language recognition.
\newblock In \emph{Proceedings of the 28th ACM International Conference on Multimedia}, pages 1366--1375, 2020.

\bibitem[Su et~al.(2024)Su, Ahmed, Lu, Pan, Bo, and Liu]{su2024roformer}
Jianlin Su, Murtadha Ahmed, Yu Lu, Shengfeng Pan, Wen Bo, and Yunfeng Liu.
\newblock Roformer: Enhanced transformer with rotary position embedding.
\newblock \emph{Neurocomputing}, 568:\penalty0 127063, 2024.

\bibitem[Tunga et~al.(2021)Tunga, Nuthalapati, and Wachs]{tunga2021pose-based}
A Tunga, S~V Nuthalapati, and J Wachs.
\newblock Pose-based sign language recognition using gcn and bert.
\newblock In \emph{Proceedings of the IEEE/CVF Winter Conference on Applications of Computer Vision}, pages 31--40, 2021.

\bibitem[Uthus et~al.(2024{\natexlab{a}})Uthus, Tanzer, and Georg]{uthus2024youtube}
Dave Uthus, Garrett Tanzer, and Manfred Georg.
\newblock Youtube-asl: A large-scale, open-domain american sign language-english parallel corpus.
\newblock \emph{Advances in Neural Information Processing Systems}, 36, 2024{\natexlab{a}}.

\bibitem[Uthus et~al.(2024{\natexlab{b}})Uthus, Tanzer, and Georg]{uthus2024youtubeasl}
D Uthus, G Tanzer, and M Georg.
\newblock Youtube-asl: A large-scale, open-domain american sign language-english parallel corpus.
\newblock \emph{Advances in Neural Information Processing Systems}, 36, 2024{\natexlab{b}}.

\bibitem[Vaia(2024)]{csl_history_grammar_vaia}
Vaia.
\newblock Chinese sign language: History, grammar.
\newblock \url{https://www.vaia.com/en-us/explanations/chinese/chinese-grammar/chinese-sign-language/}, 2024.
\newblock Accessed: November 20, 2024.

\bibitem[von Agris and Kraiss(2010)]{vonagris2010signum}
U von Agris and K~F Kraiss.
\newblock Signum database: Video corpus for signer-independent continuous sign language recognition.
\newblock In \emph{4th Workshop on the Representation and Processing of Sign Languages: Corpora and Sign Language Technologies}, pages 243--246, 2010.

\bibitem[Voskou et~al.(2021)Voskou, Panousis, Kosmopoulos, et~al.]{voskou2021stochastic}
A Voskou, K~P Panousis, D Kosmopoulos, et~al.
\newblock Stochastic transformer networks with linear competing units: Application to end-to-end sl translation.
\newblock In \emph{Proceedings of the IEEE/CVF International Conference on Computer Vision}, pages 11946--11955, 2021.

\bibitem[Wang et~al.(2016)Wang, Chai, Hong, et~al.]{wang2016isolated}
H Wang, X Chai, X Hong, et~al.
\newblock Isolated sign language recognition with grassmann covariance matrices.
\newblock \emph{ACM Transactions on Accessible Computing (TACCESS)}, 8\penalty0 (4):\penalty0 1--21, 2016.

\bibitem[wikiHow Contributors(2024)]{types_of_sign_language_wikihow}
wikiHow Contributors.
\newblock What are the types of sign language? asl, bsl, \& more.
\newblock \url{https://www.wikihow.com/Types-of-Sign-Language}, 2024.
\newblock Accessed: November 20, 2024.

\bibitem[Wong et~al.()Wong, Camgoz, and Bowden]{wong2024sign2gptleveraginglargelanguage}
Ryan Wong, Necati~Cihan Camgoz, and Richard Bowden.
\newblock Sign2gpt: Leveraging large language models for gloss-free sign language translation.
\newblock In \emph{The Twelfth International Conference on Learning Representations}.

\bibitem[Wong et~al.(2024)Wong, Camgoz, and Bowden]{wong2024sign2gpt}
R Wong, N~C Camgoz, and R Bowden.
\newblock Sign2gpt: Leveraging large language models for gloss-free sign language translation.
\newblock \emph{arXiv preprint arXiv:2405.04164}, 2024.

\bibitem[WorldAtlas(2024)]{languages_in_china_worldatlas}
WorldAtlas.
\newblock What languages are spoken in china?
\newblock \url{https://www.worldatlas.com/articles/what-languages-are-spoken-in-china.html}, 2024.
\newblock Accessed: November 20, 2024.

\bibitem[Yang et~al.(2024)Yang, Yang, Hui, Zheng, Yu, Zhou, Li, Li, Liu, Huang, et~al.]{yang2024qwen2}
An Yang, Baosong Yang, Binyuan Hui, Bo Zheng, Bowen Yu, Chang Zhou, Chengpeng Li, Chengyuan Li, Dayiheng Liu, Fei Huang, et~al.
\newblock Qwen2 technical report.
\newblock \emph{arXiv preprint arXiv:2407.10671}, 2024.

\bibitem[Yao et~al.(2023)Yao, Zhou, Feng, Hu, Zhou, and Li]{yao2023sign}
Huijie Yao, Wengang Zhou, Hao Feng, Hezhen Hu, Hao Zhou, and Houqiang Li.
\newblock Sign language translation with iterative prototype.
\newblock In \emph{Proceedings of the IEEE/CVF International Conference on Computer Vision}, pages 15592--15601, 2023.

\bibitem[Ye et~al.()Ye, Jiao, Wang, Tu, and Xiong]{ye2023cross}
Jinhui Ye, Wenxiang Jiao, Xing Wang, Zhaopeng Tu, and Hui Xiong.
\newblock Cross-modality data augmentation for end-to-end sign language translation.
\newblock In \emph{The 2023 Conference on Empirical Methods in Natural Language Processing}.

\bibitem[Ye et~al.(2024)Ye, Wang, Jiao, Liang, and Xiong]{ye2024improving}
Jinhui Ye, Xing Wang, Wenxiang Jiao, Junwei Liang, and Hui Xiong.
\newblock Improving gloss-free sign language translation by reducing representation density.
\newblock \emph{arXiv preprint arXiv:2405.14312}, 2024.

\bibitem[Yin et~al.(2023)Yin, Zhong, Tang, et~al.]{yin2023gloss}
A Yin, T Zhong, L Tang, et~al.
\newblock Gloss attention for gloss-free sign language translation.
\newblock In \emph{Proceedings of the IEEE/CVF Conference on Computer Vision and Pattern Recognition}, pages 2551--2562, 2023.

\bibitem[Yin and Read(2020)]{yin2020better}
Kayo Yin and Jesse Read.
\newblock Better sign language translation with stmc-transformer.
\newblock In \emph{Proceedings of the 28th International Conference on Computational Linguistics}, pages 5975--5989. International Committee on Computational Linguistics, 2020.

\bibitem[Zhang et~al.()Zhang, M{\"u}ller, and Sennrich]{zhang2023sltunet}
Biao Zhang, Mathias M{\"u}ller, and Rico Sennrich.
\newblock Sltunet: A simple unified model for sign language translation.
\newblock In \emph{The Eleventh International Conference on Learning Representations}.

\bibitem[Zhang et~al.(2023)Zhang, Li, and Bing]{zhang2023video}
Hang Zhang, Xin Li, and Lidong Bing.
\newblock Video-llama: An instruction-tuned audio-visual language model for video understanding.
\newblock In \emph{Proceedings of the 2023 Conference on Empirical Methods in Natural Language Processing: System Demonstrations}, pages 543--553, 2023.

\bibitem[Zhang et~al.(2016)Zhang, Zhou, Xie, et~al.]{zhang2016chinese}
J Zhang, W Zhou, C Xie, et~al.
\newblock Chinese sign language recognition with adaptive hmm.
\newblock In \emph{2016 IEEE International Conference on Multimedia and Expo (ICME)}, pages 1--6. IEEE, 2016.

\bibitem[Zhang et~al.(2024)Zhang, Wu, Li, Li, Ma, Liu, and Li]{zhang2024video}
Yuanhan Zhang, Jinming Wu, Wei Li, Bo Li, Zejun Ma, Ziwei Liu, and Chunyuan Li.
\newblock Video instruction tuning with synthetic data.
\newblock \emph{arXiv preprint arXiv:2410.02713}, 2024.

\bibitem[Zhao et~al.(2023)Zhao, Wu, He, and Huang]{zhao2023svit}
Bo Zhao, Boya Wu, Muyang He, and Tiejun Huang.
\newblock Svit: Scaling up visual instruction tuning.
\newblock \emph{arXiv preprint arXiv:2307.04087}, 2023.

\bibitem[Zhao et~al.(2021)Zhao, Qi, Zhou, et~al.]{zhao2021conditional}
J Zhao, W Qi, W Zhou, et~al.
\newblock Conditional sentence generation and cross-modal reranking for sign language translation.
\newblock In \emph{IEEE Transactions on Multimedia}, pages 2662--2672, 2021.

\bibitem[Zhao et~al.(2024)Zhao, Zhang, Fu, et~al.]{zhao2024conditional}
R Zhao, L Zhang, B Fu, et~al.
\newblock Conditional variational autoencoder for sign language translation with cross-modal alignment.
\newblock In \emph{Proceedings of the AAAI Conference on Artificial Intelligence}, pages 19643--19651, 2024.

\bibitem[Zheng et~al.(2023)Zheng, Wang, Tan, et~al.]{zheng2023cvt-slr}
J Zheng, Y Wang, C Tan, et~al.
\newblock Cvt-slr: Contrastive visual-textual transformation for sign language recognition with variational alignment.
\newblock In \emph{Proceedings of the IEEE/CVF Conference on Computer Vision and Pattern Recognition}, pages 23141--23150, 2023.

\bibitem[Zhou et~al.(2023{\natexlab{a}})Zhou, Chen, Clap{\'e}s, Wan, Liang, Escalera, Lei, and Zhang]{zhou2023gloss}
Benjia Zhou, Zhigang Chen, Albert Clap{\'e}s, Jun Wan, Yanyan Liang, Sergio Escalera, Zhen Lei, and Du Zhang.
\newblock Gloss-free sign language translation: Improving from visual-language pretraining.
\newblock In \emph{Proceedings of the IEEE/CVF International Conference on Computer Vision}, pages 20871--20881, 2023{\natexlab{a}}.

\bibitem[Zhou et~al.(2023{\natexlab{b}})Zhou, Chen, Clapés, et~al.]{zhou2023gloss-free}
B Zhou, Z Chen, A Clapés, et~al.
\newblock Gloss-free sign language translation: Improving from visual-language pretraining.
\newblock In \emph{Proceedings of the IEEE/CVF International Conference on Computer Vision}, pages 20871--20881, 2023{\natexlab{b}}.

\bibitem[Zhou et~al.(2021{\natexlab{a}})Zhou, Zhou, Qi, et~al.]{zhou2021improving}
Hao Zhou, Wengang Zhou, Weizhen Qi, et~al.
\newblock Improving sign language translation with monolingual data by sign back-translation.
\newblock In \emph{Proceedings of the IEEE/CVF Conference on Computer Vision and Pattern Recognition}, pages 1316--1325, 2021{\natexlab{a}}.

\bibitem[Zhou et~al.(2021{\natexlab{b}})Zhou, Zhou, Zhou, and Li]{STMC_MM}
Hao Zhou, Wengang Zhou, Yun Zhou, and Houqiang Li.
\newblock Spatial-temporal multi-cue network for sign language recognition and translation.
\newblock \emph{IEEE Transactions on Multimedia}, 2021{\natexlab{b}}.

\bibitem[Zhou et~al.(2021b)Zhou, Zhou, Zhou, et~al.]{zhou2021spatial-temporal}
H Zhou, W Zhou, Y Zhou, et~al.
\newblock Spatial-temporal multi-cue network for sign language recognition and translation.
\newblock \emph{IEEE Transactions on Multimedia}, 24:\penalty0 768--779, 2021b.

\bibitem[Zhu et~al.(2023)Zhu, Chen, Shen, et~al.]{zhu2023minigpt-4}
D Zhu, J Chen, X Shen, et~al.
\newblock Minigpt-4: Enhancing vision-language understanding with advanced large language models.
\newblock \emph{arXiv preprint arXiv:2304.10592}, 2023.

\bibitem[Zuo and Mak(2024)]{zuo2024improving}
R Zuo and B Mak.
\newblock Improving continuous sign language recognition with consistency constraints and signer removal.
\newblock \emph{ACM Transactions on Multimedia Computing, Communications and Applications}, 20\penalty0 (6):\penalty0 1--25, 2024.

\bibitem[Zuo et~al.(2023)Zuo, Wei, and Mak]{zuo2023natural}
R Zuo, F Wei, and B Mak.
\newblock Natural language-assisted sign language recognition.
\newblock In \emph{Proceedings of the IEEE/CVF Conference on Computer Vision and Pattern Recognition}, pages 14890--14900, 2023.

\end{thebibliography}
}

\clearpage
\setcounter{page}{1}
\maketitlesupplementary

\renewcommand\thesection{\Alph{section}}
\renewcommand*{\theHsection}{appedix.\thesection}
\setcounter{section}{0}

\renewcommand{\thefigure}{\Alph{figure}}
\setcounter{figure}{0}
\renewcommand{\thetable}{\Alph{table}}
\setcounter{table}{0}

This supplementary material provides more qualitative results (\cref{sec:appendix:qualitative}), dataset details (\cref{sec:appendix:dataset}), 
experiment details (\cref{sec:appendix:experiment}), 
and more discussion on limitations (\cref{sec:appendix:limitations}) and broader impact (\cref{sec:appendix:social}).\\
\vspace{-8pt}

\section{Qualitative Results}
\label{sec:appendix:qualitative}
We provide more qualitative results generated by our model on the CSL-Daily~\cite{zhou2021improving} and Phoenix-2014T~\cite{camgoz2018neural} datasets, as detailed in Tab.~\ref{tab:a} and Tab.~\ref{tab:b}, respectively.
We randomly sample several translation outputs along with their corresponding spoken language references for analysis.
The results indicate that our LLaVA-SLT model is capable of generating sentences that accurately preserve the semantic essence of the source sentences, despite exhibiting structural variations.

\section{Dataset Details}
\label{sec:appendix:dataset}

\myparagraph{CSL-400h dataset.}
In the creation of the CSL-400h dataset, meticulous attention was devoted to curating a large-scale high-quality sign language video collection, which is critical to advance research in gloss-free sign language translation. This dataset encompasses a broad spectrum of sign language expressions, captured through a rigorously structured data collection process.

The dataset compilation involved \textit{\textbf{participants}} including both professional and non-professional hard-of-hearing signers to ensure a comprehensive representation of sign language usage. The professional group included three experts in sign language linguistics and several seasoned Deaf signers, while the non-professional cohort comprised individuals from diverse demographic backgrounds, including different ages, genders, occupations, and educational levels. This diverse participant pool helped capture a wide array of signing habits and styles, essential for the richness of the CSL-400h dataset.
Data collection was conducted in a \textit{\textbf{controlled environment}}, carefully designed to simulate real-life interaction scenarios (see Tab.~\ref{tab:appendix:csl400h}). This controlled setting was outfitted with high-resolution cameras and optimal lighting conditions to ensure the clarity and accuracy of the recorded sign language videos, thereby facilitating precise analysis and application in sign language translation models.

\begin{figure*}[tbp]
    \centering
    \includegraphics[width=1\linewidth]{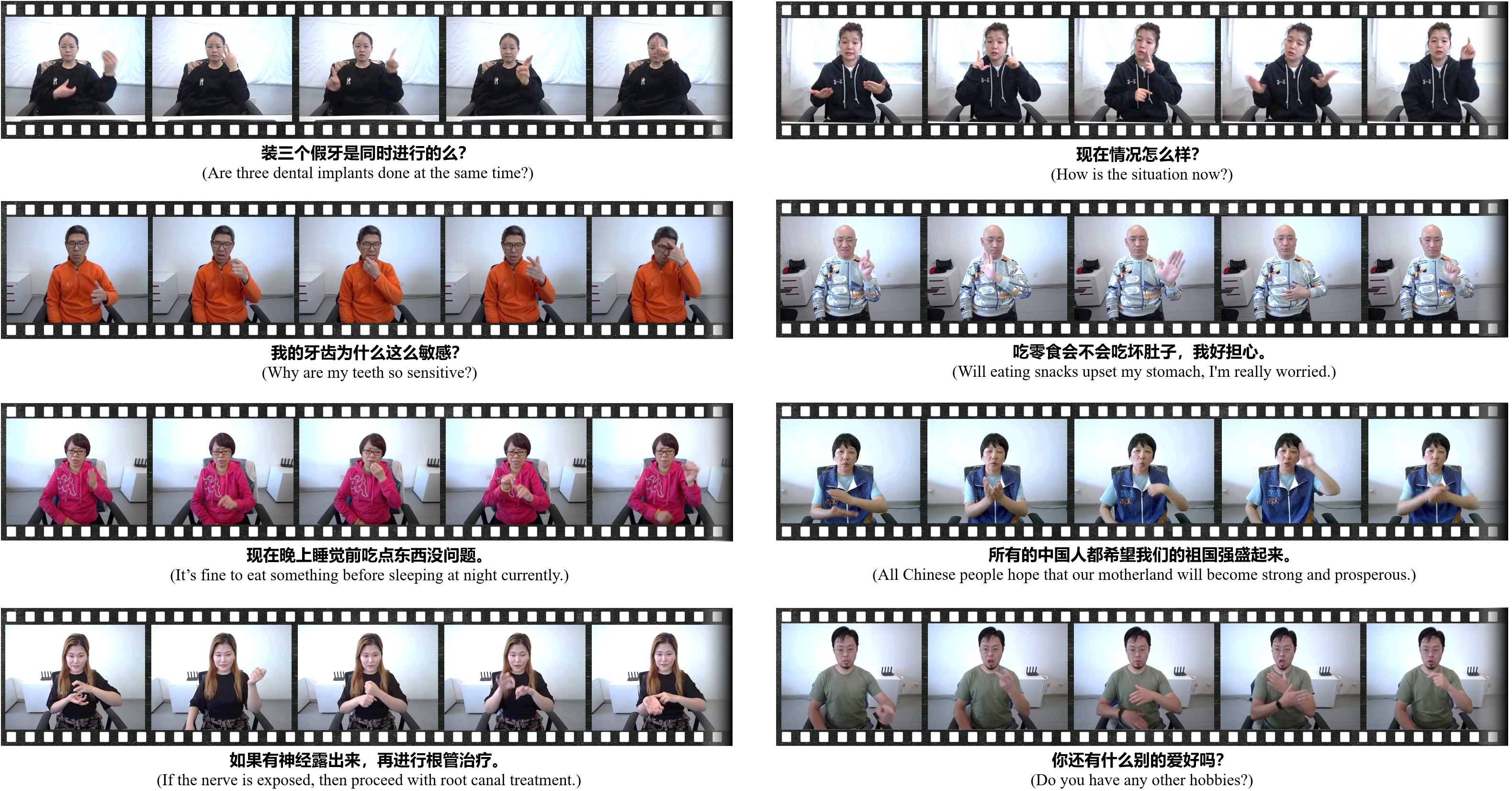}
    \caption{\textbf{CSL-400h dataset gallery.} We showcase several sign videos and their corresponding source texts in the dataset. }
    \label{fig:appendix:gallery}
    \vspace{-5pt}
\end{figure*}

As a results, the CSL-400h dataset is extensive, covering a wide range of scenarios, from professional interactions in medical, work, and educational settings to everyday situations such as entertainment, family communication, and shopping. The detailed statistics can be found in Tab.~\ref{tab:appendix:csl400h}. It comprises 120,535 clips shorter than 5 seconds and 143,926 clips longer than 5 seconds, with an average clip duration of approximately 5.56 seconds. The average sentence length within these clips is 12.54 characters, underscoring the dataset’s comprehensive nature. Some examples from the CSL-400h dataset are illustrated in Fig.~\ref{fig:appendix:gallery}, showcasing the diversity and applicability of the data in various contexts.

\begin{table}[tbp]
\centering
\scalebox{0.76}{
  \begin{tabular}{p{1.5cm} p{8.5cm} }
    \toprule
    \multicolumn{2}{l}{\bf CSL Qualitative Results}  \\
    \midrule
    Reference & \begin{CJK}{UTF8}{gbsn} 他把自己装得聪明。\end{CJK} \\
              &\textbf{(He makes himself look smart.)} \\
    Ours & \begin{CJK}{UTF8}{gbsn} 他把自己打扮成一个聪明人。\end{CJK} \\
              &\textbf{(He makes himself look like a smart person.)} \\
    \midrule
    Reference & \begin{CJK}{UTF8}{gbsn}我喜欢冬天，下雪太美了。\end{CJK} \\
              &\textbf{(I like winter; the snow is so beautiful.)} \\

    Ours & \begin{CJK}{UTF8}{gbsn} 我喜欢冬天，雪景太美了。\end{CJK} \\
              &\textbf{(I like winter; the snowy scenery is so beautiful.)} \\
    \midrule
    Reference & \begin{CJK}{UTF8}{gbsn} 我的老板对我满意，我很高兴。\end{CJK} \\
              &\textbf{(My boss is satisfied with me; I am very happy.)} \\

    Ours & \begin{CJK}{UTF8}{gbsn} 老板对我的工作很满意，我感到很高兴。\end{CJK} \\
              &\textbf{(Boss is satisfied with my work; I feel very happy.)} \\
    \midrule
    Reference & \begin{CJK}{UTF8}{gbsn} 有些小孩子很害怕输液，一直大哭。\end{CJK} \\
              &\textbf{(Some children are very scared of IV, and cry incessantly.)} \\

    Ours & \begin{CJK}{UTF8}{gbsn} 小孩子害怕输液，哭闹不止。\end{CJK} \\
              &\textbf{(Children are scared of IV, and couldn't stop crying.)} \\
    \bottomrule
  \end{tabular}
}
\caption{\textbf{Qualitative results} on CSL-Daily test set.}  
\label{tab:a}
\end{table}

\begin{table}[tb]
\centering
\scalebox{0.7}{
  \begin{tabular}{p{1.3cm} p{9.5cm} }
    \toprule
    \multicolumn{2}{l}{\bf DGS Qualitative Results}  \\
    \midrule
    Reference & \begin{CJK}{UTF8}{gbsn} am tag vor allem im norden regen.\end{CJK} \\
              &\textbf{(During the day, mainly rain in the north.)} \\
    Ours &  tagsüber vor allem regen im norden.\\
              &(\textbf{During the day, mainly rain in the north.}) \\
    \midrule
    Reference & \begin{CJK}{UTF8}{gbsn}der wind weht schwach bis mäßig aus süd bis südost.\end{CJK} \\
              &\textbf{(The wind blows weakly to moderately from south to southeast.)} \\

    Ours &  der wind weht mäßig von süd nach südost.\\
              &(\textbf{The wind is blowing moderately from south to southeast.}) \\

    \midrule
    Reference & \begin{CJK}{UTF8}{gbsn} sonst aber funkeln oft die sterne oder es bildet sich nebel.\end{CJK} \\
              &\textbf{(Otherwise, the stars often twinkle or fog forms.)} \\

    Ours &  die sterne funkeln oder es wird sich nebel bilden. \\
              &(\textbf{The stars will twinkle or fog will form.}) \\
        \midrule
    Reference & \begin{CJK}{UTF8}{gbsn} am samstag überquert uns ein regengebiet von west nach ost.
\end{CJK} \\
              &\textbf{(On Saturday a rain area will cross from west to east.)} \\

    Ours & ein regengebiet wird am samstag von westen nach osten ziehen. \\
              &(\textbf{A rain area will move from west to east on Saturday.}) \\
    \bottomrule
  \end{tabular}
}
\caption{\textbf{Qualitative results} on Phoenix-2014T test set.}  
\label{tab:b}
\end{table} 

\begin{table}[tbp]
\centering
\scalebox{0.8}{
    \begin{tabular}{@{}lcccccc@{}}
        \toprule
        \bf Scene & \bf Texts & \bf Videos & \bf Chars & \bf Duration \\ \midrule
        Work & 11,038 & 52,512 & 673,821 & 81.75h \\
        Campus & 5,987  & 29,704 & 367,549 & 44.05h \\
        Shopping & 1,005  & 5,129  & 60,856  & 7.19h  \\
        Family & 8,635  & 41,926 & 518,936 & 62.13h \\
        Entertainment & 7,430  & 37,752 & 460,877 & 55.97h \\
        Medical & 7,393  & 54,601 & 666,603 & 76.59h \\
        Others & 9,297  & 42,837 & 541,099 & 73.79h \\ \midrule
        Total & 50,785 & 264,461 & 3,296,943 & 401.47h \\ \bottomrule
    \end{tabular}
}
\caption{Detailed statistics of the CSL-400h Dataset.}  
\label{tab:appendix:csl400h}
\vspace{-5pt}
\end{table}

\myparagraph{CSL-Corpus dataset.}
To collect high-quality Chinese Sign Language (CSL) \textit{\textbf{gloss-text pairs}} data efficiently, we follow a rigorous procedure that adheres to precise guidelines. 
For the spoken language side, we employ GPT-4 to generate text sentences given the predefined scenarios, which is prompted to cover linguistic content and vocabulary as broad as possible.
For the sign language side, all vocabulary is required to align with the \textit{National Universal Sign Language Dictionary.} In instances where spoken vocabulary lacks a corresponding sign language entry, we adopt the closest equivalent phrases as recommended by sign language experts.
The textual Chinese sign language (gloss sequence) annotation was performed to mirror the natural syntax of sign language, distinct from spoken Chinese. This approach guarantees that these translations are not mere replicas of spoken language but conform to the linguistic structures familiar to the deaf community.
Concerning the length and accuracy of the textual sign language sentences, we ensure that the sign language word count does not exceed their corresponding spoken language version. This constraint helps keep the translations concise yet true to the original meanings without altering. Special attention is paid to the placement of question words, negations, adjectives, and numerals, aligning them with the syntactic norms of sign language. 
Each gloss-text pair is meticulously developed to be clear and functional for the deaf community, meeting these strict standards to support precise and natural communication. Several examples are shown in Tab.~\ref{tab:appendix:gloss}.

Furthermore, to enhance the domain knowledge of LLMs, we compile \textit{\textbf{additional corpora}}, sourced from online webpages, academic papers, and books.
The topics of these materials are centered around CSL, including general knowledge, research papers, textbooks, and dictionaries. A detailed list of these sources is provided in Tab.~\ref{tab:appendix:corpus}.
We standardize the electronic versions of these materials into TXT format for training, ensuring the inclusion of high-quality textual content and tabular data.

\begin{table}[tbp]
\centering
\scalebox{0.73}{
  \begin{tabular}{p{1.cm} p{9.3cm} }
    \toprule
    \multicolumn{2}{l}{\bf CSL-Corpus: Gloss-Text Pairs}  \\
    \midrule
    text & \begin{CJK}{UTF8}{gbsn} 生活中怎样防止皮肤过敏？\end{CJK} \\
              &\textbf{(How to prevent skin allergies in daily life?)} \\
    gloss & \begin{CJK}{UTF8}{gbsn} 生活/防止/皮肤/病/怎样/?\end{CJK} \\
              &\textbf{(daily life/prevent/skin/illness/how/?)} \\
    \midrule
    text & \begin{CJK}{UTF8}{gbsn}他几乎每周都会来两三次。\end{CJK} \\
              &\textbf{(He comes almost two or three times a week.)} \\

    gloss & \begin{CJK}{UTF8}{gbsn} 他/基本/每周/来/二/三 \end{CJK} \\
              &\textbf{(he/almost/every week/come/two/three)} \\
    \midrule
    text & \begin{CJK}{UTF8}{gbsn} 累死我了，真的不想再逛了。\end{CJK} \\
              &\textbf{(I'm exhausted, really don't want to wander around anymore.)} \\

    gloss & \begin{CJK}{UTF8}{gbsn} 我/累/想/逛街/不 \end{CJK} \\
              &\textbf{(I/exhausted/want to/wander around/no)} \\
    \midrule
    text & \begin{CJK}{UTF8}{gbsn} 我最喜欢的还是墨西哥煎饼。\end{CJK} \\
              &\textbf{(My favorite is still Mexican pancakes.)} \\

    gloss & \begin{CJK}{UTF8}{gbsn} 我/喜欢/墨西哥/煎/饼  \end{CJK} \\
              &\textbf{(I/like/Mexican/fried/cake)} \\
    \bottomrule
  \end{tabular}
}
\caption{\textbf{Gloss-text pairs gallery.} We showcase several gloss-text pairs in the CSL-Corpus dataset.}  
\label{tab:appendix:gloss}
\end{table}

\begin{table}[ht]
\centering
\scalebox{0.79}{
\begin{tabular}{p{9cm} l}
\toprule
\textbf{Title} & \textbf{Type} \\ \midrule
\begin{CJK}{UTF8}{gbsn} 专访上海大学倪兰教授：语言学与手语识别技术的融合突破，解锁交流障碍 - GAIR live~\cite{ni_lan_interview_2024} \end{CJK} & Web \\
\begin{CJK}{UTF8}{gbsn} 中国手语 - 维基百科，自由的百科全书~\cite{chinese_sign_language_wikipedia} \end{CJK} & Web \\
\begin{CJK}{UTF8}{gbsn} 中国手语/Chinese Sign Language - Non-Mandarin Chinese - Chinese-Forums~\cite{chinese_sign_language_forums} \end{CJK} & Web \\
\begin{CJK}{UTF8}{gbsn} 手语 - 维基百科，自由的百科全书~\cite{sign_language_wikipedia} \end{CJK} & Web \\
\begin{CJK}{UTF8}{gbsn} 手语（聋人交流方式） - 百度百科~\cite{sign_language_baidu} \end{CJK} & Web \\
\begin{CJK}{UTF8}{gbsn} 手语翻译 - 学习手语 - 手语图解 - 手语查询~\cite{sign_language_learning_bmcx} \end{CJK} & Web \\
\begin{CJK}{UTF8}{gbsn} What Languages Are Spoken in China? - WorldAtlas~\cite{languages_in_china_worldatlas} \end{CJK} & Web \\
\begin{CJK}{UTF8}{gbsn} What Are the Types of Sign Language? ASL, BSL, \& More~\cite{types_of_sign_language_wikihow} \end{CJK} & Web \\
\begin{CJK}{UTF8}{gbsn} Making Themselves Heard: Chinese Sign Language \& Deaf China Online~\cite{deaflepuff_chinese_sign_language} \end{CJK} & Web \\
\begin{CJK}{UTF8}{gbsn} Is American Sign Language different from Chinese Sign Language? - Quora~\cite{asl_vs_csl_quora} \end{CJK} & Web \\
\begin{CJK}{UTF8}{gbsn} IDSL: Unleashing dreams and protecting rights of the hearing-impaired - CGTN~\cite{idsl_cgtn_dreams_rights} \end{CJK} & Web \\
\begin{CJK}{UTF8}{gbsn} IDSL: China's sign language history dates back to Tang Dynasty period - CGTN~\cite{idsl_cgtn_history} \end{CJK} & Web \\
\begin{CJK}{UTF8}{gbsn} Deaf, not Dumb: Chinese Sign Language - Sinosplice~\cite{idsl_cgtn_history} \end{CJK} & Web \\
\begin{CJK}{UTF8}{gbsn} CSL, The Revolutionary Chinese Sign Language - Sign Language Blogs~\cite{csl_sign_language_blogs} \end{CJK} & Web \\
\begin{CJK}{UTF8}{gbsn} Connaissez-Vous la Langue des Signes Chinoise ? Un Guide Complet sur 中国手语~\cite{csl_french_ltl_school} \end{CJK} & Web \\
\begin{CJK}{UTF8}{gbsn} Chinese Sign Language: History, Grammar - Vaia~\cite{csl_history_grammar_vaia} \end{CJK} & Web \\
\begin{CJK}{UTF8}{gbsn} Chinese Sign Language - Wikipedia~\cite{csl_wikipedia_english} \end{CJK} & Web \\
\begin{CJK}{UTF8}{gbsn}中国手语中的“指点”手势研究~\cite{LvHuaihua2017} \end{CJK} & Paper \\
\begin{CJK}{UTF8}{gbsn}中国手语语料库高频词初步分析及标注探讨~\cite{LiuXueda2022} \end{CJK} & Paper \\
\begin{CJK}{UTF8}{gbsn}中国手语音系学中的音节结构研究~\cite{WangXiaoxia2020} \end{CJK} & Paper \\
\begin{CJK}{UTF8}{gbsn}中国手语一般疑问句中疑问手控标记研究~\cite{LinHao2018} \end{CJK} & Paper \\
\begin{CJK}{UTF8}{gbsn}中国手语相同音系参数下的词义拓展和相关性构建~\cite{WenXu2022} \end{CJK} & Paper \\
\begin{CJK}{UTF8}{gbsn}中国手语基本手势语义知识组织研究~\cite{ZhangYanqun2024} \end{CJK} & Paper \\
\begin{CJK}{UTF8}{gbsn}中国手语定中短语的语序~\cite{LvHuaihuaWangHongying2017} \end{CJK} & Paper \\
\begin{CJK}{UTF8}{gbsn}论中国手语的分类词谓语~\cite{YaoDengfeng2018} \end{CJK} & Paper \\
\begin{CJK}{UTF8}{gbsn}The Influence of Chinese Characters on Chinese Sign Language~\cite{RenTianyu2024} \end{CJK} & Paper \\
\begin{CJK}{UTF8}{gbsn}Chinese Sign Language Recognition Based on Surface Electromyography and Motion Information~\cite{LiWenyu2023} \end{CJK} & Paper \\
\begin{CJK}{UTF8}{gbsn}中国手语动词研究~\cite{NiLan2015} \end{CJK} & Book \\
\begin{CJK}{UTF8}{gbsn}上海手语音系~\cite{ZhangJisheng2019} \end{CJK} & Book \\
\begin{CJK}{UTF8}{gbsn}中国手语（上下修订版）~\cite{ZhongguoShouyu} \end{CJK} & Book \\
\begin{CJK}{UTF8}{gbsn}国家通用手语词典~\cite{GuoTongyongShouyu} \end{CJK} & Book \\
\begin{CJK}{UTF8}{gbsn}手语基础教程~\cite{ShouyuJichu} \end{CJK} & Book \\
\begin{CJK}{UTF8}{gbsn}手语会话~\cite{ShouyuHuihua} \end{CJK} & Book \\ \hline
\end{tabular}
}
\caption{Detailed sources of additional corpora in the CSL-Corpus dataset.}
\label{tab:appendix:corpus}
\end{table}

\section{Experiment Details}
\label{sec:appendix:experiment}

\myparagraph{Benchmark setup.}
The experimental framework was rigorously designed to ensure the robustness and reproducibility of our findings. We detail the benchmark datasets employed in our study as follows:

\begin{itemize} 
\setlength\itemsep{0em}

    \item \textbf{CSL-Daily}: This dataset focuses on Chinese sign language, also annotated with glosses and translations. It includes 18,401 training instances, 1,077 validation instances, and 1,176 test instances.

    \item \textbf{Phoenix-2014T}: An extension of the Phoenix-2014 dataset, which is dedicated to German sign language, specifically targeting weather forecast content. It comprises 7,096 training instances, 519 validation instances, and 642 test instances.

\end{itemize} 

\myparagraph{Input processing.}
For alignment with previous methods~\cite{wong2024sign2gptleveraginglargelanguage,zhou2023gloss}, our approach incorporates an initial image input processing phase. Specifically, for our model with a resolution of 
$336^2$, input sequences are initially centrally cropped as squares and resized to $384^2$, and randomly/centrally cropped to achieve a final resolution of
$336^2$ during the training/inference phases, respectively. For configurations requiring a resolution of
$224^2$, an intermediate resizing to
$256^2$ is implemented. To further enhance model robustness and generalization across diverse visual contexts, a suite of data augmentation techniques, including color jitter, frame rotations, and horizontal flips, is consistently applied across all frames within the video sequences to maintain temporal coherence.

\myparagraph{Instruction prompt templates.}
Furthermore, we provide comprehensive templates for instructional prompts that are employed during the Visual Language Tuning phase of our framework. The templates designated for translation into Chinese Sign Language (CSL) and German Sign Language (DGS) translation are illustrated in Tab.~\ref{tab:appendix:template1} and Tab.~\ref{tab:appendix:template2}, respectively.
In the training process, we fulfill the video placeholder with the output generated by the Multilayer Perceptron (MLP) connector. Concurrently, the text placeholder is replaced by the corresponding text ground truth in spoken language. To facilitate effective model training, both the proposed task prompt and the format prompt are appended at the end of the system prompt segment.

\begin{table}[tp]
\centering
\scalebox{0.99}{
  \begin{tabular}{p{1.5cm} p{6.cm} }
    \toprule
    \multicolumn{2}{l}{\bf CSL Instruction Template}  \\ 
    \midrule
    System    &You are a helpful assistant. \textbf{Use your expertise to provide the most precise translation. Answer with one single Chinese sentence.} \\ \\

    User & $<\left|video\_placeholder\right|>$ \\ \\

    Assistant & $<\left|text\_placeholder\right|>$ \\ 

    \bottomrule
  \end{tabular}
}
\caption{\textbf{Instruction template} for Chinese Sign Language (CSL) translation task. Our task and format prompts are highlighted.}  
\label{tab:appendix:template1} 
\end{table}

\begin{table}[tp]
\centering
\scalebox{0.99}{
  \begin{tabular}{p{1.5cm} p{6cm} }
    \toprule
    \multicolumn{2}{l}{\bf DGS Instruction Template}  \\ 
\midrule
    System    &You are a helpful assistant. \textbf{Use your expertise to provide the most precise translation. Answer with one single German sentence.} \\ \\

    User & $<\left|video\_placeholder\right|>$ \\  \\

    Assistant & $<\left|text\_placeholder\right|>$ \\ 

    \bottomrule
  \end{tabular}
}
\caption{\textbf{Instruction template} for German Sign Language (DGS) translation task. Our task and format prompts are highlighted.}  
\label{tab:appendix:template2} 
\end{table}

\section{Limitations}
\label{sec:appendix:limitations}
Despite the promising results demonstrated by LLaVA-SLT, several limitations need to be addressed in future work.

\myparagraph{Long-term Context.}
The current implementation of LLaVA-SLT primarily addresses short-term contexts within a single sign language sentence. Sign language, however, inherently relies on long-term dependencies and extensive contextual information that may span several sentences or entire conversations. To enhance the coherence and accuracy of translations in extended interactions, it is imperative to develop methodologies that effectively integrate long-term contextual information.

\myparagraph{Real-world Scenarios.}
The training and evaluation of our model have predominantly utilized datasets like CSL-Daily, CSL-400h and Phoenix-2014T, which are captured in controlled settings. These datasets may not adequately represent the complexities and variabilities encountered in natural environments. Factors such as variable lighting conditions, background noise, and the diversity in signer appearances can drastically affect performance. Therefore, advancing the robustness of LLaVA-SLT in real-world conditions remains a critical area for future research.

\myparagraph{Multi-turn QA Dialogue.}
Currently, LLaVA-SLT is optimized for single-turn translation tasks. However, practical applications often necessitate handling multi-turn question-answer dialogues, where maintaining contextual continuity across exchanges is essential. Developing strategies to manage these multi-turn interactions efficiently is crucial for evolving LLaVA-SLT into a more interactive and user-centric sign language AI agent.

\section{Broader Impact}
\label{sec:appendix:social}
The development of the LLaVA-SLT framework holds potential social impact, particularly for the hard-of-hearing communities. By improving the accuracy and accessibility of sign language translation, our work can enhance communication between hard-of-hearing and hearing individuals, fostering inclusivity and understanding. This advancement is crucial in various settings, such as education, where it can facilitate better learning experiences for hard-of-hearing students, and healthcare, where clear communication is vital for patient care. Moreover, our framework can be instrumental in creating more inclusive workplaces and public services, ensuring that hard-of-hearing individuals have equal opportunities and access. By open-sourcing our code and models, we aim to encourage further research and development in sign language processing, driving innovation and contributing to a more inclusive society. We believe that our work will contribute to not only bridging communication gaps but also promoting social equity.

\end{document}